%% file: main_IEEE.tex
\documentclass[conference]{IEEEtran}
\IEEEoverridecommandlockouts

\input{include_packages}           

\def\BibTeX{{\rm B\kern-.05em{\sc i\kern-.025em b}\kern-.08em
    T\kern-.1667em\lower.7ex\hbox{E}\kern-.125emX}}
    
\begin{document}
\input{include_title}              

\maketitle
\input{sections/s0_abstract}


\input{include_all_sections}     

\bibliographystyle{IEEEtran}
\bibliography{reference} 
\balance
\include{sections/s9_appendix}

\end{document}

%% file: include_packages.tex
\usepackage{cite}
\usepackage[pdftex]{graphicx}
\usepackage{amsmath}
\usepackage{amssymb}
\usepackage{amsthm}
\newtheorem{assumption}{Assumption}
\usepackage{algorithm}
\usepackage{algorithmic}
\usepackage{array}
\usepackage[caption=false,font=footnotesize]{subfig}
\usepackage{dblfloatfix}
\usepackage{url}
\usepackage[numbers]{natbib}
\usepackage{enumitem}
\usepackage{xspace}

\newcommand{\attname}{\textsc{DejaVu}\xspace}

\usepackage{xcolor}
\usepackage{hyperref}
\usepackage{balance}

\usepackage{tikz}

\usepackage{amsthm}

\usepackage{tikz}            
\newcommand{\circledlabel}[1]{%
  \tikz[baseline=(char.base)]{
    \node[circle, fill=black, text=white, inner sep=0.75pt] (char) {$\scriptstyle #1$};
  }%
}

\definecolor{myblue}{rgb}{0.0, 0.0, 1.0}
\definecolor{myblue}{rgb}{0.0, 0.0, 0.0}
\definecolor{mydarkgreen}{rgb}{0.0, 0.5, 0.0}

\newcommand\todo[1]{\textcolor{red}{#1}}

\renewcommand\todo[1]{}

\usepackage{mdframed}

\newmdenv[
  backgroundcolor=gray!10,
  linecolor=black!20,
  linewidth=0.3pt,
  roundcorner=2pt,
  skipabove=6pt,
  skipbelow=6pt
]{keyfinding}

\usepackage{soul}
\usepackage{xcolor}

%% file: include_title.tex
\title{Temporal Misalignment Attacks against Multimodal Perception in Autonomous Driving}



\author{
\IEEEauthorblockN{
Md Hasan Shahriar\IEEEauthorrefmark{1}
\thanks{This work has been accepted for publication at the IEEE Conference on Secure and Trustworthy Machine Learning (SaTML 2026). The final version will be available on IEEE Xplore.}
,
Md Mohaimin Al Barat\IEEEauthorrefmark{1},
Harshavardhan Sundar\IEEEauthorrefmark{2},
Ning Zhang\IEEEauthorrefmark{3},\\
Naren Ramakrishnan\IEEEauthorrefmark{1},
Y.~Thomas Hou\IEEEauthorrefmark{1},
Wenjing Lou\IEEEauthorrefmark{1}
}

\IEEEauthorblockA{
\IEEEauthorrefmark{1}Virginia Tech, Blacksburg, VA, USA\\
\texttt{\{hshahriar, barat, naren, thou, wjlou\}@vt.edu}
}

\IEEEauthorblockA{
\IEEEauthorrefmark{2}Amazon.com, Inc., New York, NY, USA\\
\texttt{hsundar427@gmail.com}
}

\IEEEauthorblockA{
\IEEEauthorrefmark{3}Washington University in St. Louis, St. Louis, MO, USA\\
\texttt{zhang.ning@wustl.edu}
}
}

%% file: sections/s0_abstract.tex
\begin{abstract}
Multimodal fusion (MMF) plays a critical role in the perception of autonomous driving, which primarily fuses camera and LiDAR streams for a comprehensive and efficient scene understanding. However, its strict reliance on precise temporal synchronization exposes it to new vulnerabilities. In this paper, we introduce \attname, an attack that exploits the in-vehicular network to manipulate the integrity of time and create subtle temporal misalignments, severely degrading downstream MMF-based perception tasks. 
Our comprehensive attack analysis across different models and datasets reveals the sensors' task-specific imbalanced sensitivities: object detection is overly dependent on LiDAR inputs, while object tracking is highly reliant on the camera inputs. Consequently, with a single-frame LiDAR delay, an attacker can reduce the car detection mAP by up to 88.5\%, while with a three-frame camera delay, multiple object tracking accuracy (MOTA) for car drops by 73\%. 
We further demonstrated two attack scenarios using an automotive Ethernet testbed for hardware-in-the-loop validation and the Autoware stack for end-to-end AD simulation, demonstrating the feasibility of the \attname attack and its severe impact, such as collisions and phantom braking. Our code and artifacts are publicly available at \url{https://github.com/shahriar0651/DejaVu}. 
\end{abstract}

%% file: include_all_sections.tex
\input{sections/s1_introduction}

\input{sections/s1_related_works}

\input{sections/s2_preliminaries}

\input{sections/s3_misalignment_attack}
\input{sections/s4_hil_experiments}
\input{sections/s5_model_experiments}

\input{sections/s7_discussion}

\input{sections/s8_conclusion}

%% file: sections/s1_introduction.tex
\section{Introduction}
\label{sec:introduction}


Autonomous driving (AD) is designed to navigate and interact with complex environments---a capability fundamentally reliant on a comprehensive understanding of its surroundings. The adoption of heterogeneous sensors, such as camera, LiDAR, and radar, that capture data from different modalities allows an accurate perception with enhanced accuracy and robustness~\cite{zhang2023multi}. Each individual modality has unique strengths; for example, cameras capture rich semantic details, LiDAR provides accurate depth measurements, and radar particularly excels in detecting speed, even in adverse weather conditions~\cite{marti2019review}. However, these sensors also face inherent limitations, such as a camera's sensitivity to lighting variations, LiDAR's lack of texture information, and radar's sparsity, which can compromise performance when used independently~\cite{wang2021can, de2020evaluating, engels2021automotive}.  Multimodal fusion (MMF)---the process of integrating multiple unimodal sensor data into a single and comprehensive representation---compensates for such individual sensor weaknesses, ensures accurate and robust perception, and efficient downstream tasks in AD~\cite{bramon2012multi, cheng2023fusion}. 
MMF remains a fundamental research challenge, primarily due to the heterogeneity across sensing modalities, including differences in data formats and spatial~\cite{huang2022multi, feng2020deep}.

Another central aspect of this heterogeneity is \textit{temporal}: in practice, multimodal sensors operate at inherently different sampling rates and are naturally asynchronous. Cameras typically capture frames at 30--60\,Hz, LiDAR at 10--20\,Hz, and radar at yet different rates. Each sensor samples the environment independently, so messages arrive at the fusion node at different times and with different temporal spacing. As a result, a camera frame captured at a given instant often has no LiDAR or radar measurement captured at exactly the same moment. Before fusion can take place, the system must therefore decide \emph{which} observations from each modality correspond to the same physical instant---a problem known as \textit{temporal alignment}. Getting this pairing right is critical. Fusing observations from different times, known as misalignments, yields semantically inconsistent representations, degrading perception accuracy and leading to object misdetection, localization errors, and scene misinterpretation. These inaccuracies propagate downstream to control, maneuver planning, and safety interventions, ultimately compromising vehicle reliability and safety~\cite{chen2024end}. 
However, the robustness of temporal alignment has received comparatively limited attention~\cite{kuhse2024sync}, despite being a fundamental enabler of MMF in AD. Moreover, it remains a complex problem due to the following challenges.
\begin{figure*}[t]
\centering
\includegraphics[width=0.99\textwidth]{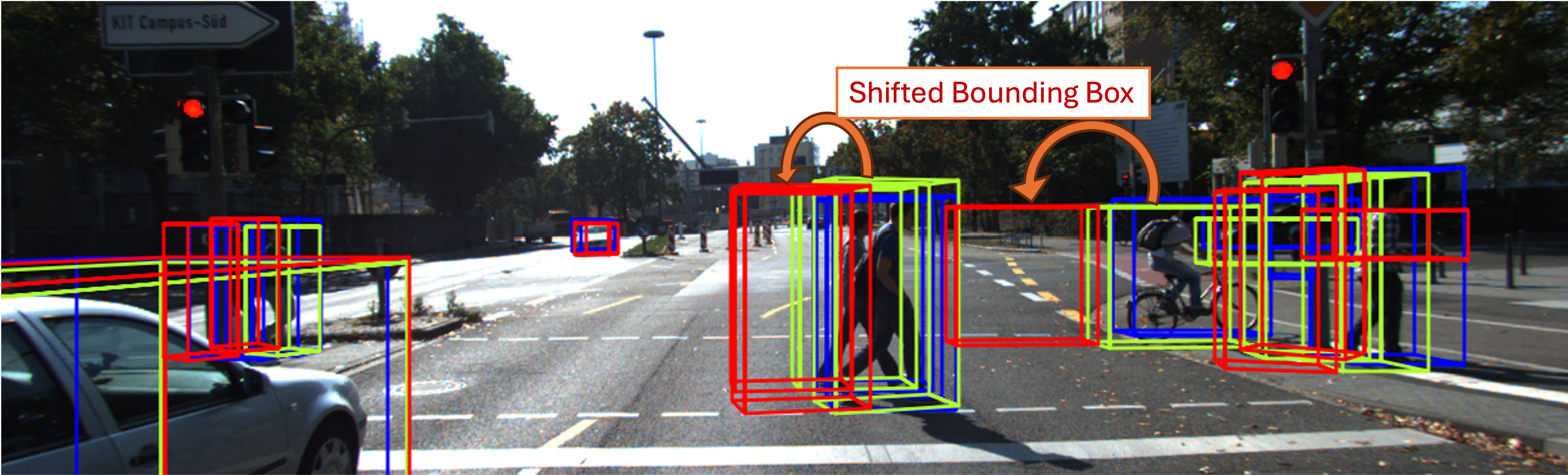}
    \label{fig:attack-camera}
\caption{
Visualization of the impact of the \attname attack on a camera--LiDAR fusion-based perception model.
Ground-truth boxes are shown in \textcolor{blue}{blue}, benign predictions in \textcolor{green!60!black}{green}, and predictions under \attname in \textcolor{red}{red}.
MMF-based fused predictions are overlaid on the current camera frame.
While benign predictions closely track the ground truth, delaying the LiDAR stream by five frames under \attname induces cross-sensor temporal misalignment, shifting predicted bounding boxes toward objects’ past locations and causing some objects to be missed.
This behavior suggests an overreliance on LiDAR inputs relative to the camera stream.
A corresponding LiDAR-plane visualization is provided in the Appendix (Fig.~\ref{fig:example_dejavu_attack_v3}).
}

\vspace{-5pt}
\label{fig:example_dejavu_attack_v2}
\end{figure*}

\textbf{Clock synchronization.} To ensure temporally aligned perception in AD, sensor data from all the modalities must be timestamped relative to a common global clock with minimal drift. 
In practice, the electronic controller units (ECUs) hosting the sensors must be tightly synchronized---often at sub-microsecond precision required by AUTOSAR~\cite{autosar2020timesync})---to enable accurate fusion and downstream reasoning. To achieve that, automotive Ethernet (AE) equipped with Time-Sensitive Networking (TSN) has become the de facto backbone of modern in-vehicle networks. 
TSN leverages the generalized Precision Time Protocol (gPTP, IEEE 802.1AS) to maintain global clock synchronization among distributed ECUs~\cite{stanton2018distributing, deng2022survey}. Consequently, a core security assumption is: \circledlabel{A_1} \textbf{\textit{The gPTP-based synchronization infrastructure is secure and maintains consistent global time across all ECUs.}} 
However, despite its widespread adoption in AE, gPTP was primarily designed to provide deterministic time synchronization---not to operate under adversarial conditions. Hence, gPTP lacks built-in security mechanisms, particularly cryptographic authentication, leaving it vulnerable to attacks such as grandmaster spoofing, delay injection, replay, and false time advertisement~\cite{shi2023ms, finkenzeller2024ptpsec, finkenzeller2025securing}. Such attacks can compromise the temporal alignment of sensor fusion, undermining the integrity of MMF-based perception, without directly manipulating the sensor ECU itself. 

\textbf{Timestamp Integrity.~} Even when sensors are nominally synchronized, they often capture data at slightly different times due to variations in sampling rates caused by inherent hardware limitations~\cite{kuhse2024sync}. Consequently, the timestamps attached to sensor messages become critical in multimodal fusion pipelines, as they serve as the primary reference for aligning asynchronous data streams by identifying the most temporally proximal pairs from buffered queues. This reliance gives rise to a fundamental security assumption: 
\circledlabel{A_2} \textbf{\textit{Sensor ECUs are trusted entities that always provide accurate timestamps.}} 
However, this assumption can be violated. In vehicular systems, sensor ECUs may be compromised through remote exploits of unpatched software vulnerabilities~\cite{checkoway2011comprehensive, foster2015fast, miller2015remote}, insecure Over-The-Air (OTA) updates~\cite{yeasmin2021multi, ghosal2022secure}, or direct physical access via the OBD-II port~\cite{avatefipour2018state, eiza2017driving, checkoway2011comprehensive}. Once compromised, an attacker-controlled ECU can inject messages containing legitimate sensor data but with forged timestamps. Since the fusion ECU uses timestamps for temporal alignment, this can lead to selecting data pairs that appear temporally aligned but are semantically misaligned, thereby degrading the integrity of the sensor fusion process.


\textbf{Middleware Integrity.~} 
Sensor data exchange in production-grade AD (such as Autoware\footnote{https://github.com/autowarefoundation/autoware}) is often facilitated by robotic middleware frameworks, such as robot operating system (ROS\footnote{https://github.com/ros2/ros2}), where the data distribution service (DDS) serves as the underlying communication backbone. DDS enables a real-time and scalable publish-subscribe communication model for data sharing among distributed ECUs, making it a widely adopted solution for managing the high-bandwidth, low-latency communication demands of multimodal perception and control pipelines. Consequently, a third foundational assumption arises: \circledlabel{A_3} \textbf{\textit{The ROS-based DDS infrastructure is secure and ensures the integrity, authenticity, and freshness of shared data.}} However, the design of ROS prioritizes performance and scalability over robust security guarantees. Since ROS does not enforce strong authentication, encryption, or source verification by default, ROS is vulnerable to a wide range of network-level and application-level attacks, including message spoofing, replay, and impersonation attacks~\cite{deng2022security}. As a result, an attacker with access to the ROS communication graph, for example, can impersonate legitimate nodes (e.g., a LiDAR publisher) and can publish fabricated sensor messages with targeted timestamps (i.e., replay attacks), which can be considered for fusion and mislead downstream perception tasks.

Moreover, prior work has demonstrated that existing MMF-based perception systems in AD lack \textit{temporal robustness}---the ability to maintain reliable outputs in the presence of temporal inconsistencies---making them vulnerable even to benign, non-malicious misalignments~\cite{kuhse2024sync, yeong2021sensor, huck2011precise}. In particular, perception pipelines have been shown to degrade significantly under small delays in just one modality. 
Existing studies, however, are limited in both scope and threat modeling: they primarily focus on random, system-induced delays and evaluate only a narrow subset of perception tasks---most commonly 3D object detection, without realizing them on any end-to-end AD software stacks.
To investigate the true fragility of MMF-based perception under adversarial conditions, we present a comprehensive study of temporal misalignment attacks, which we term \attname. These attacks are realized by violating one or more of the core trust assumptions.
Unlike prior work~\cite{finkenzeller2025sensor}, we target both 3D object detection and multi-object tracking (MOT), and evaluate the effects of different delay distributions across multiple perception models and datasets. 
Fig.~\ref{fig:example_dejavu_attack_v2} illustrates how temporal delays due to \attname attacks in one of the modalities can degrade the performance of the MVXNet model~\cite{sindagi2019mvx}---a MMF-based 3D object detection model, potentially leading to unsafe driving conditions in AD. In summary, we make the following key contributions:

\begin{itemize}[leftmargin=*]
    \item We propose \attname, a temporal misalignment attack against MMF in AD that exploits vulnerabilities of in-vehicle networks and the fragility of multimodal fusion by selectively delaying sensor streams to disrupt perception in safety-critical tasks.
    \item We conduct a comprehensive empirical evaluation of \attname on state-of-the-art 3D object detection models (MVXNet~\cite{sindagi2019mvx}, BEVFusion~\cite{liu2023bevfusion}) using the KITTI~\cite{geiger2012we} and nuScenes~\cite{caesar2020nuscenes} datasets, and a multi-object tracking model (MMF-JDT~\cite{wang2024multi}) under various misalignment scenarios using the KITTI~\cite{geiger2012we} dataset. Our findings reveal distinct modality-specific vulnerabilities: object detectors are highly sensitive to LiDAR delays, while the tracking model is significantly impacted by camera timing disruptions. A single-frame LiDAR delay reduces 3D detection mAP by up to 88.5\%, and a three-frame camera delay drops multiple object tracking accuracy (MOTA) by 73\%.
    \item To further validate our findings in a realistic autonomous driving setting, we built an automotive Ethernet testbed that models the sensor data acquisition and fusion pipeline. Using this platform, we implemented the \attname attack by violating both \circledlabel{A_1} and \circledlabel{A_2}, demonstrating its feasibility in a hardware-in-the-loop environment. Additionally, to assess the end-to-end consequences beyond perception---specifically on planning and control---we integrated \attname into Autoware, a production-grade, full-stack autonomous driving simulator, by breaking \circledlabel{A_3}. In both environments, our experiments show that \attname is highly practical and can result in severe safety violations, including direct collisions and phantom braking events.

\end{itemize}

The remainder of the paper is organized as follows. 
The remainder of the paper is organized as follows. Section~\ref{sec:relatedworks} reviews related work. Section~\ref{sec:ad-system-model} formalizes the AD system model. Section~\ref{sec:attacking-mmf} presents the threat model and introduces the \attname attack. Section~\ref{sec:exp-hil} describes the realization of the attack on an HIL testbed. Section~\ref{sec:exp-setting-att} details the datasets and evaluation settings. Section~\ref{sec:eval-results} reports the evaluation results. Section~\ref{sec:discussion} discusses the findings and potential defenses. Section~\ref{sec:conclusion} concludes the paper.

%% file: sections/s1_related_works.tex
\section{Related Work}
\label{sec:relatedworks}







The fundamental research direction has been in the direction of spoofing sensors from a single modality, such as LiDAR~\cite{cao2023you, jin2024phantomlidar, hallyburton2022security} and camera~\cite{wang2024revisiting} through different means of physical perturbation. Moreover, advanced attacks have demonstrated to even compromise multiple modalities together~\cite{zhu2024malicious, cao2021invisible, bagdasaryan2024adversarial}. 
These sensor spoofing attacks demonstrate that simply having multiple sensors is not sufficient. With carefully constructed inputs, an adversary can simultaneously mislead camera and LiDAR sensors, defeating the very redundancy meant to ensure safety.

Unlike sensor spoofing attacks, time delay attacks have not been extensively studied within AD. They have created significant attention in other cyber-physical systems (CPS) domains, such as power systems~\cite{xiahou2020robust,shangguan2022resilient}, wireless networks~\cite{song2007attack}, unmanned aerial vehicles (UAVs)~\cite{zhai2023etd,zhai2023hotd, shi2023ms}, and time-sensitive networks~\cite{luo2023impact}. Software timing interference is also exploited to cause system destabilization in CPS~\cite{li2021chronos}. 
Moreover, multimodal temporal misalignment in sensor fusion has been shown to degrade the accuracy of simultaneous localization and mapping (SLAM)~\cite{li2022timing}, which was limited only to the fusion between IMU and camera data. 
Contrary to the existing works, we comprehensively study the impact of a temporal misalignment attack on task-agnostic MMF-based perception. 


In the realm of AD security, various defense mechanisms have been proposed to counteract sensor spoofing~\cite{sato2025realism} and multimodal fusion attacks. These defenses can be broadly categorized into spatiotemporal consistency checks, specification-aware recovery strategies, and hardware-based techniques~\cite{gao2021autonomous}. 
PercepGuard~\cite{man2023person} detects misclassification attacks by enforcing consistency between object tracks and class labels, but it does not examine the temporal validity of sensor readings and thus cannot detect replayed LiDAR scans whose semantic trajectories remain plausible. Connecting the Dots~\cite{li2020connecting} employs class‐specific autoencoders to uncover context violations introduced by adversarial perturbations, yet time‐shifted data aligns perfectly with learned scene co‐occurrence statistics and evades its checks. PhyScout~\cite{xu2024physcout} formalizes cross‐modal conflict detection to identify gross spoofing, but subtle timestamp manipulations within the synchronizer’s tolerance window introduce no overt spatial or modality discrepancies. These approaches overlook \emph{timestamp integrity} and data freshness as a security property. 

%% file: sections/s2_preliminaries.tex
\section{Autonomous Driving System Model}
\label{sec:ad-system-model}

This section provides a formal end-to-end model of AD systems, covering data collection, temporal alignment, and MMF-based perception. 

\subsection{Data Collection and Temporal References}
\label{sec:data-collection}

\begin{figure}
    \centering
    \includegraphics[width=0.99\linewidth]{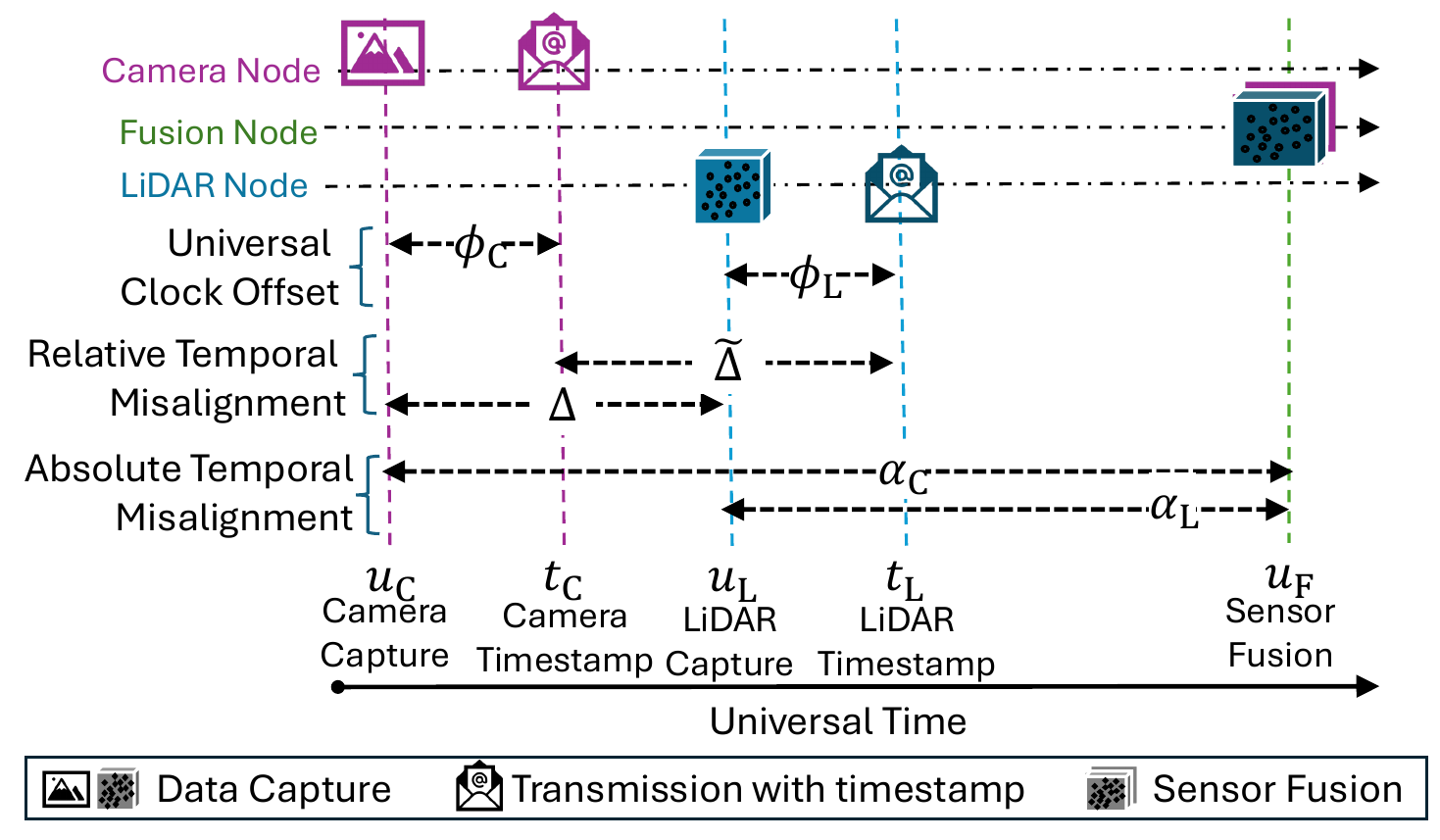}
    \caption{Illustration of temporal events in MMF-based perception for autonomous driving systems. Under benign operating conditions, the UCO ($\phi_S$), the reported and actual RTMs ($\tilde{\Delta}$ and $\Delta$), and the S-ATM values ($\alpha_S$) are all approximately negligible, up to the system’s end-to-end pipeline latency.}
    \vspace{-5pt}
    \label{fig:temporal-events}
\end{figure}

Modern AD systems rely on heterogeneous sensor arrays to perceive the environment. We focus on camera ($S_C$) and LiDAR ($S_L$) as representative modalities, though the model extends to additional sensors (e.g., radar, IMU). Each sensor $S \in \{S_C, S_L\}$ transmits measurements over in-vehicle networks to a fusion node and produces a sequence of messages $m_S^{(i)} = \big(x_S^{(i)},\; t_S^{(i)}\big)$ indexed by $i \in \mathbb{N}$.
A critical modeling detail is the distinction between two time references associated with each measurement. The \emph{universal capture time} $u_S^{(i)}$ denotes the true physical time when a data sample $x_S^{(i)}$ is captured in the world. As an estimation of $u_S^{(i)}$, the sensor attaches a \emph{timestamp} $t_S^{(i)}$ using its local clock to the message header. 
The time reference $t_S^{(i)}$ is used by the fusion node for sorting, buffering, and temporal alignment of the message queue. Figure~\ref{fig:temporal-events} illustrates such temporal events at different sensor nodes.  For analysis purposes, we quantify the discrepancy between these two time references by the \emph{universal clock offset} (UCO) of sensor $S$:
\begin{equation}
\label{eqn:uco}
    \phi_S^{(i)} = \left|t_S^{(i)} - u_S^{(i)}\right|.
\end{equation}
In practice, sensors operate under nominal conditions where clock synchronization and minimal processing delays keep the universal clock offset small (i.e., $\phi_S^{(i)} \approx 0$). We formalize benign operation via the following assumption.
\begin{assumption}[Timestamp fidelity]
\label{assum:timestamp-fidelity}
For each sensor message, the local timestamp approximates the universal capture time, i.e., $t_S^{(i)} \approx u_S^{(i)}$ (equivalently, $\phi_S^{(i)} \approx 0$).
\end{assumption}


\begin{figure*}[!t]
    \centering
    \includegraphics[width=0.995\linewidth]{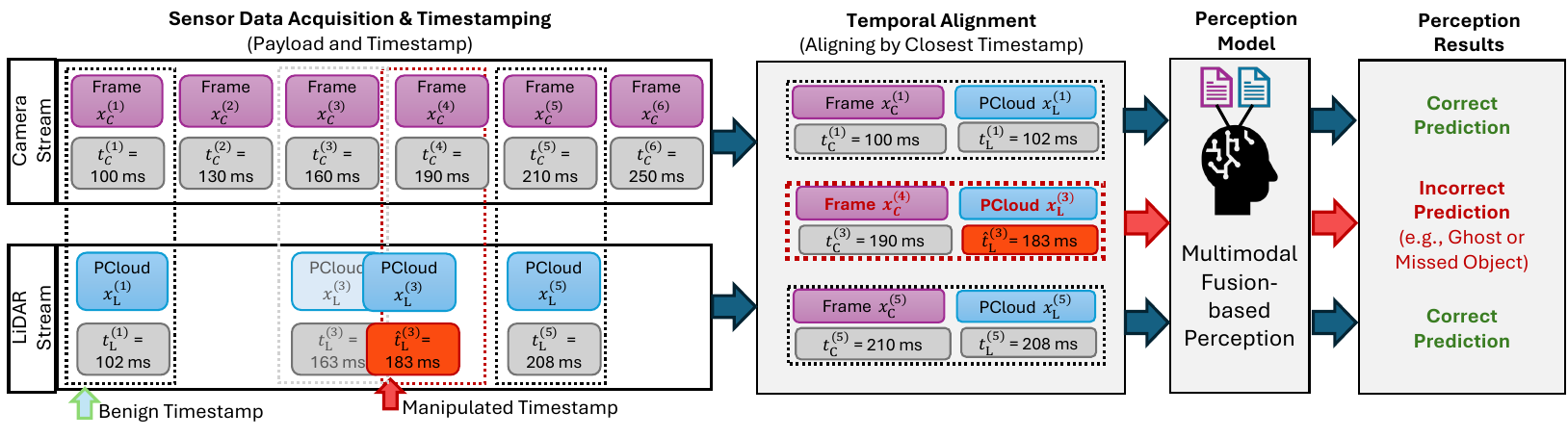}
    \vspace{-10pt}
    \caption{\textbf{Overview of the \attname Attack.} Camera and LiDAR sensors produce asynchronous data streams $\{x_C^{(i)}\}$ and $\{x_L^{(j)}\}$. The adversary maliciously modifies a LiDAR packet by manipulating its timestamp $t_L^{(3)}$ (i.e., 183 ms instead of 163 ms) while keeping the data payload benign. During approximate temporal synchronization, the corresponding LiDAR point cloud is incorrectly paired with a temporally proximate camera frame (e.g., $(x_C^{(4)}, x_L^{(3)})$). This temporally misaligned pair is then fused by the perception model, leading to degraded predictions such as ghost objects or missed detections.}
\vspace{-5pt}
  \label{fig:dejavu_attack}    
\end{figure*}

\subsection{Temporal Alignment}
\label{sec:temporal-alignment}


Due to asynchronous sampling and transmission, temporal alignment is a critical step for MMF. Each modality maintains a finite FIFO buffer $\mathcal{Q}_S$ containing the most recent $N$ messages, ordered by $t_S^{(\cdot)}$ (the only time reference available to the system). The fusion node employs a \emph{temporal alignment} mechanism to identify temporally proximal cross-modal pairs that correspond to approximately the same physical instant.


A widely used policy is \textit{approximate-time synchronization}, where the fusion node pairs messages from different modalities by comparing their header timestamps and accepting pairs whose timestamp difference is within a tolerance $\tau$ (the \emph{slop}). This mechanism is implemented in production AD stacks such as ROS~2's \texttt{ApproximateTimeSynchronizer}\footnote{\texttt{TimeSynchronizer} and \texttt{ApproximateTimeSynchronizer} are commonly used message filtering utilities in ROS~2 that align multiple sensor message streams based on their timestamps. While \texttt{TimeSynchronizer} performs strict timestamp matching, \texttt{ApproximateTimeSynchronizer} allows messages with slight temporal differences---within a specified tolerance window---to be synchronized.} and similar middleware. Concretely, for each unpaired camera message $m_C^{(i)} \in \mathcal{Q}_C$, the aligner finds the LiDAR index:

\begin{equation}
j^\star(i) = \arg\min_{k:\, m_L^{(k)} \in \mathcal{Q}_L} \big| t_C^{(i)} - t_L^{(k)} \big|.    
\end{equation}
If $\big|t_C^{(i)} - t_L^{(j^\star(i))}\big| \le \tau$, the fusion node selects the pair $(i, j^\star(i))$ for fusion and remove both messages from their buffers. 
Symmetrically apply the rule for unpaired LiDAR messages $m_L^{(j)}$. 
For an selected pair $(i,j)$, we quantify temporal alignment in two complementary ways:
\begin{itemize}[leftmargin=*]
    \item {Reported relative temporal misalignment (R-RTM):} $\tilde{\Delta}^{(i,j)} = \lvert t_C^{(i)} - t_L^{(j)} \rvert$, which must satisfy $\tilde{\Delta}^{(i,j)} \le \tau$ for the aligner to accept the pair.
    \item {Semantic relative temporal misalignment (S-RTM):} $\Delta^{(i,j)} = \lvert u_C^{(i)} - u_L^{(j)} \rvert$, which captures the true capture-time separation between the paired observations.
\end{itemize}

Under benign synchronization and Assumption~\ref{assum:timestamp-fidelity}, the aligned pairs satisfy $|t_C^{(i)} - t_L^{(j)}| \ll \tau$ and holds ($t_S^{(i)} \approx  u_S^{(i)}$), this implies that the paired payloads are also close in capture time, so fusion inputs are semantically aligned and the reported and semantic RTM coincide, i.e., $\tilde{\Delta}^{(i,j)}=\Delta^{(i,j)}$. 
This alignment policy bases pairing exclusively on reported timestamps $t_S^{(.)}$ and queue state, \emph{not} on message content. Thus, while the above-mentioned assumptions are effective under benign conditions, it creates a vulnerability: an adversary who can manipulate timestamps can force the alignment mechanism to pair semantically inconsistent observations.


\subsection{Multimodal Fusion-based Perception}
\label{sec:fusion}

As each heterogeneous sensor generates data samples with unique formats and dimensions, modality-specific unimodal feature encoders $E_S$ convert raw data $x_S^{(i)}$ to intermediate representations $f_S^{(i)} = E_S(x_S^{(i)})$. These encoders map raw data to a common feature format, enabling various fusion techniques such as concatenation, attention mechanisms, or tensor-based fusion~\cite{xiang2023multi}. For each aligned pair $(i, j)$ selected by the alignment mechanism, the fusion module computes a fused representation $z^{(i,j)}$ using a fusion operator $\mathcal{F}_{\theta}$:
$$
z^{(i,j)} = \mathcal{F}_{\theta}\big(f_C^{(i)}, f_L^{(j)}\big).
$$
We associate each selected pair $(i,j)$ with a \emph{universal fusion time} $u_F^{(i,j)}$, when the fusion actually happens, and the system perceives the environment. For a fused pair $(i,j)$, we define the {semantic absolute temporal misalignment (S-ATM)} $\alpha_S^{(i,j)}$ as the age of each data sample relative to the fusion event:
\begin{equation}
\alpha_C^{(i,j)} = \lvert u_F^{(i,j)} - u_C^{(i)} \rvert,
\qquad
\alpha_L^{(i,j)} = \lvert u_F^{(i,j)} - u_L^{(j)} \rvert.
\end{equation}
With a reliable communication infrastructure, this leads to the following assumptions:
\begin{assumption}[Fusion-time Coherence]
\label{assum:fusion-time-coherence}
For an selected pair $(i,j)$, the universal fusion time $u_F^{(i,j)}$ is close to the capture time of each sensor, so that $u_F^{(i,j)} \approx u_C^{(i)} \approx u_L^{(j)}$ and $\alpha_C^{(i,j)} \approx \alpha_L^{(i,j)} \approx 0$.
\end{assumption}

The corresponding perception head $\mathcal{H}_{\theta}$ then produces the final prediction: $\mathcal{Y}^{(i,j)} = \mathcal{H}_{\theta}\big(z^{(i,j)}\big)$. 
The perception output $\mathcal{Y}^{(i,j)}$ typically includes object detections, semantic segmentation, or tracking states, and is then consumed by downstream planning and control modules.




%% file: sections/s3_misalignment_attack.tex

\section{Temporal Misalignment Attack : \attname}
\label{sec:attacking-mmf}

In this section, we formalize the \attname attack, where an adversary maliciously introduces temporal misalignment by manipulating the local timestamps used for temporal alignment. The attack exploits the timestamp-based pairing mechanism, which can keep pairs aligned in the reported timestamp space while making them semantically misaligned in universal capture time, degrading fusion quality. Figure~\ref{fig:dejavu_attack} illustrates an example of \attname attack. 



\subsection{Threat Model and Attack Surface}
\label{sec:threat-model}

\subsubsection{Adversary Objective and Scope}
The adversary's goal is to coerce the temporal alignment mechanism to emit \emph{semantically inconsistent} (misaligned) pairs so that the fused embedding $z^{(i,j)}$ combines observations from different physical times, degrading downstream performance (e.g., missed or spurious detections, localization errors). The adversary is \emph{timing-only}: they do not modify payloads $x_S^{(i)}$ or model parameters, ensuring the attack remains stealthy against content-based detectors.
The attack breaks benign operation via two linked steps. First, the adversary injects a universal clock offset (UCO) so that Assumption~\ref{assum:timestamp-fidelity} ($t_S^{(i)} \approx u_S^{(i)}$) no longer holds and $\phi_S^{(i)}$ becomes large. Second, because the aligner pairs messages using timestamps only, the poisoned queues can yield pairs that satisfy the reported tolerance window ($\tilde{\Delta}^{(i,j)}\le \tau$) yet violate Assumption~\ref{assum:fusion-time-coherence} by producing large $\Delta^{(i,j)}$ and/or large $\alpha_S^{(i,j)}$.

\subsubsection{Attacker Capabilities}
\label{sec:attacker-capabilities}

The adversary can gain access to the in-vehicle network through one or more realistic entry points~\cite{de2024systematic}. 
\textit{Physical Access:} The attacker connects to the vehicle's OBD-II port or directly to Ethernet/CAN interfaces, either through malicious maintenance personnel or via physical compromise~\cite{checkoway2011comprehensive}.
\textit{Remote Exploitation:} The attacker exploits vulnerabilities in externally exposed interfaces, such as telematics units, infotainment systems, or over-the-air (OTA) update mechanisms~\cite{checkoway2011comprehensive, miller2015remote, ghosal2022secure}.
\textit{Supply Chain Attacks:} The attacker implants malicious code or hardware during manufacturing, allowing persistent access post-deployment~\cite{yang2024robustness}.

Once access is established, the attacker can monitor, intercept, and inject messages on the in-vehicle AE backbone and associated sub-networks. We assume the attacker operates under any of the following three distinct capabilities:

\textit{\circledlabel{C_1} Disruption of Clock Synchronization.~} With this capability, the adversary targets the clock synchronization mechanism in AE---specifically, the gPTP (IEEE 802.1AS). Rather than altering timestamps directly, the attacker compromises the synchronization process itself, thereby inducing \textit{actual} temporal misalignment between sensor streams. This can be accomplished by impersonating the grandmaster clock or tampering with synchronization messages via selective delay, replay, or man-in-the-middle attacks~\cite{decusatis2019impact, alghamdi2021precision, annessi2018encryption}. Although individual timestamps are not directly manipulated, they no longer correspond to a consistent global time due to induced drift, impairing the performance of perception tasks.

\textit{\circledlabel{C_2} Manipulation of Timestamp Integrity.~} In this scenario, the attacker is much powerful and has direct control over the ECU. Under this attack, the attacker preserves the actual flow of data but alters the timestamps embedded in transmitted packets~\cite{de2024systematic}. This results in \textit{seemingly} temporally misaligned messages, even though the underlying data remain \textit{actually} temporally aligned. Consequently, the fusion node introduces semantic misalignment during the alignment process. 
This capability is practically plausible because many ROS2 deployments are not configured with authentication-by-default~\cite{deng2022security}, and ECUs frequently run third-party or legacy software that enlarges the attack surface~\cite{checkoway2011comprehensive, foster2015fast, miller2015remote, yeasmin2021multi, ghosal2022secure}.

\textit{\circledlabel{C_3} Impersonation of a Legitimate Node in ROS2.~} In this scenario, the attacker is a participant in the ROS2 network. As the default ROS2 implementation lacks built-in security mechanisms---a configuration commonly adopted in industry-grade AD software stacks, including Autoware---the attacker can instantiate a malicious node that impersonates a legitimate sensor publisher. By subscribing to target sensing topics\footnote{Representative topic names from Autoware are \texttt{/sensing/camera/traffic\_light/image\_raw}, \texttt{/sensing/lidar/top/pointcloud\_raw}, etc.}, the attacker gains access to the data stream and records historical sensor messages. The malicious node then re-publishes previously captured messages with updated, genuine-looking timestamps, potentially while the original publisher is still active. By repeatedly injecting such delayed-but-valid messages, the attacker can pollute the input queue of time-based synchronizers, increasing the likelihood that a forged message is selected during the fusion process.

\subsection{Attack Formalization}
\label{sec:attack-formalization}



For any message $m_S^{(k)}=(x_S^{(k)},t_S^{(k)})$, the adversary injects a timestamp perturbation $\delta_S^{(k)}$ so that, while the payload $x_S^{(k)}$ remains unchanged, the local timestamp used by the aligner becomes inconsistent with the universal capture time:
\[
\hat{t}_S^{(k)} = t_S^{(k)} + \delta_S^{(k)}, \qquad \phi_S^{(k)} = \left| \hat{t}_S^{(k)}-u_S^{(k)}\right| \gg 0.
\]
For concreteness, consider a unimodal \attname where only the camera stream is compromised, i.e., $\hat{t}_C^{(i)} \neq t_C^{(i)}$ while $\hat{t}_L^{(j)} = t_L^{(j)}$. The temporal alignment mechanism then computes pairings using the (potentially compromised) timestamps:
\[
\hat{j}^\star(i) = \arg\min_{k:\, m_L^{(k)} \in \mathcal{Q}_L} \big| \hat{t}_C^{(i)} - t_L^{(k)} \big|,
\]
and emits the pair $(i, \hat{j}^\star(i))$ if $\big|\hat{t}_C^{(i)} - t_L^{(\hat{j}^\star(i))}\big| \le \tau$.
A stealthy adversary ensures the reported RTM stays within tolerance (i.e., $\tilde{\Delta}^{(i,\hat{j}^\star(i))}\le\tau$), so each emitted pair appears valid to the aligner. Meanwhile, by increasing the UCO and poisoning the timestamp-ordered queues, the attacker can inflate the actual S-RTM $\Delta^{(i,\hat{j}^\star(i))}$ and/or the S-ATM values $\alpha_S^{(i,\hat{j}^\star(i))}$, causing the fused representation to combine observations from distinct physical instants. 
These semantically inconsistent fused inputs yield corrupted perception outputs $\hat{\mathcal{Y}}^{(i,\hat{j}^\star(i))} \neq \mathcal{Y}^{(i,j)}$ (e.g., missed detections, false positives, localization drift), which can propagate to planning and control. We quantify these effects empirically in Section~\ref{sec:exp-setting-att} and Section~\ref{sec:eval-results}.

\subsection{Attack Strategies and Delay Distributions}
The adversary can craft different attack strategies by controlling the injected timestamp perturbation schedule $\delta_S^{(k)}$ (and hence the induced $\phi_S^{(k)}$), which can vary over time depending on the attacker's intent. Based on the attacker's capabilities, the adversary can compromise either a single modality (unimodal \attname, Uni-\attname) or multiple modalities simultaneously (multimodal \attname, Mul-\attname). 
Moreover, based on the distribution of the delay $\delta_S^{(k)}$, we consider two representative types of attack as summarized in Table~\ref{tab:temporal_attacks} and described as follows:

\begin{table}[!t]
    \caption{Two Types of \attname Attack Strategy}
    \label{tab:temporal_attacks}
    \centering
    \renewcommand{\arraystretch}{1.3}
    \begin{tabular}{|l|c|c|}
        \hline
        \textbf{Attack Name} & \textbf{Attack Type} &         \textbf{Timestamp Perturbation} $\delta_S^{(k)}$ \\
        \hline
        Constant Delay & Constant & $\delta_S^{(k)} = \Delta_{\text{lag}}$ (constant offset) \\
        \hline
        Random Delay & Random & $\delta_S^{(k)} \sim \text{Uniform}(-\Delta_{\max}, \Delta_{\max})$ \\
        \hline
    \end{tabular}
\end{table}

\subsubsection{Constant Delay Attack} This strategy applies a fixed timestamp offset $\delta_S^{(k)} = \Delta_{\text{lag}}$ to all messages from a target sensor within a time window. The attacker can achieve this through capabilities \circledlabel{C_1}--\circledlabel{C_3}, such as creating clock desynchronization, tampering with timestamps, or replaying messages with forged timestamps. 
\textbf{Impact:} The alignment mechanism pairs messages with a consistent temporal lag, causing the fusion model to receive temporally consistent but shifted data. This leads to delayed (misaligned) perception, manifesting as missing or shifted bounding boxes in object detection, where real-time perception is essential.

\subsubsection{Random Delay Attack}
In this strategy, each message experiences a different timestamp perturbation, randomly sampled from $\delta_S^{(k)} \sim \text{Uniform}(-\Delta_{\max}, \Delta_{\max})$. This strategy not only disrupts real-time requirements but also disrupts the temporal sequence of sensor data, which is crucial for object tracking. 
\textbf{Impact:} The fusion system struggles to maintain proper ordering of sensor inputs, leading to degraded perception accuracy. This causes erratic behavior in time-sensitive sequential applications, particularly for object tracking and autonomous navigation.

%% file: sections/s4_hil_experiments.tex
\section{Realization of \attname}
\label{sec:exp-hil}

This section realizes \attname on a hardware-in-the-loop (HIL) Automotive Ethernet testbed and empirically characterizes the induced temporal misalignment on this HIL testbed.
\subsection{Automotive Ethernet HIL Testbed}
\label{sec:hil}

\subsubsection{Testbed description}
\begin{figure}[t]
    \centering
    \includegraphics[width=0.495\textwidth]{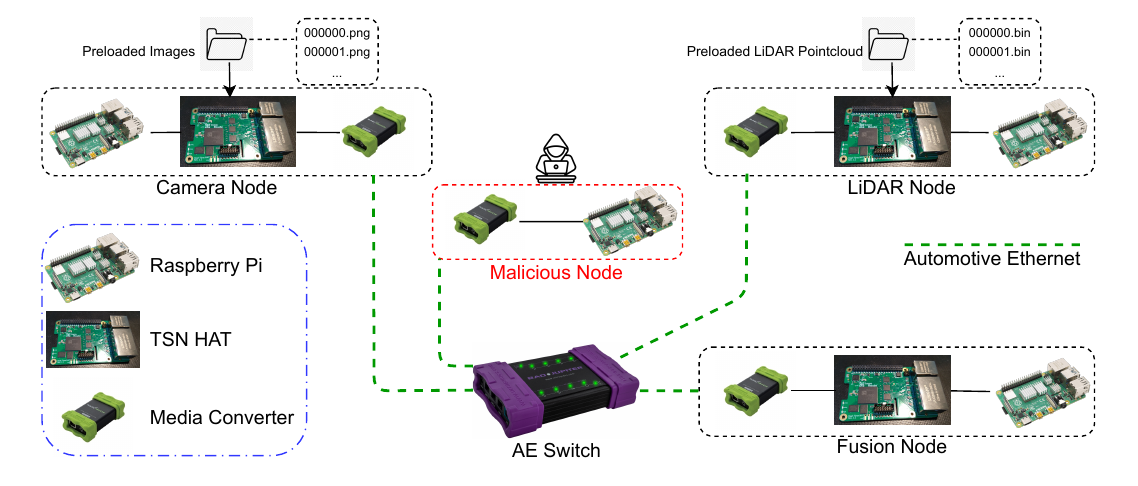}
    \caption{Schematic diagram of hardware-in-the-loop testbed.}
    \label{fig:testbed_hardware}
\end{figure}

We evaluate \attname on a hardware-in-the-loop Automotive Ethernet testbed that mirrors the system model of Section~\ref{sec:ad-system-model} and serves as a proxy for an in-vehicle AD perception network. As illustrated in Fig.~\ref{fig:testbed_hardware}, the testbed comprises four Raspberry Pi nodes: a camera node, a LiDAR node, a fusion node, and a fourth node that is within the network but not part of the MMF pipeline. For reproducibility and controlled conditions, the camera and LiDAR nodes replay the \textit{KITTI Tracking Dataset} instead of live sensors. Each node is equipped with a RealTime TSN HAT and a media converter; the nodes are interconnected via an AE switch to provide TSN and AE connectivity.

ROS~2 over DDS is used for message distribution. The camera and LiDAR nodes publish ROS~2 messages (images and point clouds) at a fixed rate (default $\approx 10$\,Hz). Each message carries (i) the sensor payload and (ii) a header timestamp set from the node's wall clock at publish time. All nodes run in a single PTP domain and are synchronized with \texttt{ptpd}. The fusion node subscribes to both topics and runs ROS~2's \texttt{ApproximateTimeSynchronizer}: it maintains bounded per-topic queues and emits a camera--LiDAR tuple when the header timestamps of the two messages fall within a configurable tolerance (the slop $\tau=0.10$\,s). The fusion node then runs MMF-based perception on each emitted pair.

This setup instantiates the pipeline of Section~\ref{sec:ad-system-model}: the header timestamp is the local timestamp $t_S^{(i)}$, and the universal capture time at which the payload was captured (or replayed) is $u_S^{(i)}$. Under benign PTP synchronization, Assumption~\ref{assum:timestamp-fidelity} holds, the UCO $\phi_S^{(i)}$ stays near zero, and the synchronizer's reported R-RTM $\tilde{\Delta}^{(i,j)}$ is a faithful proxy for the actual S-RTM $\Delta^{(i,j)}$. In the remainder of this section, we realize \attname under different attacker capabilities and empirically characterize the induced temporal misalignment (UCO, R-RTM, S-RTM, and S-ATM) on this HIL testbed.

\begin{figure}[t]
\centering
\subfloat[Local timestamps under attack]{
    \includegraphics[width=0.24\textwidth]{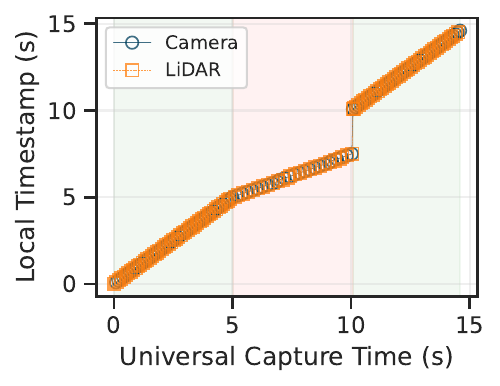}
    \label{fig:ptp-attack-c1-a}
}
\subfloat[Temporal misalignment metrics]{
    \includegraphics[width=0.24\textwidth]{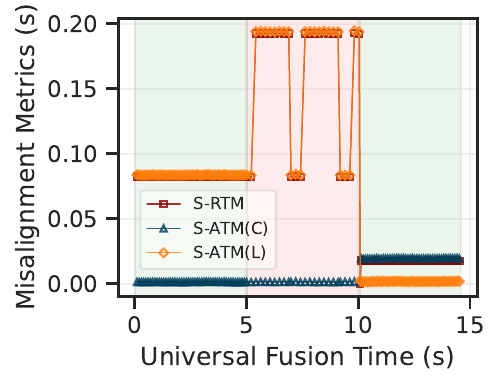}
    \label{fig:ptp-attack-c1-b}
}
\caption{Impact of the PTP-based \attname attack on timestamped sensor streams and temporal alignment in the HIL testbed. Following an initial benign phase, the attack is launched at $t{=}5$,s, during which the adversary repeatedly steps the GM clock backward ($\delta_S{=}0.20$,s every $0.22$,s). Panel (a) illustrates the resulting distortion in the camera and LiDAR local (header) timestamps. Panel (b) reports the corresponding alignment metrics for selected sensor pairs: while remaining within the acceptable slope threshold ($<0.10$) under benign conditions, the S-RTM and S-ATM values exhibit pronounced spikes during the attack, indicating semantically misaligned fusion inputs.}
\label{fig:ptp-attack-c1}
\end{figure}

We realize the \attname attack under capability \circledlabel{C_1} (disruption of clock synchronization) as described below. An instantiation under capability \circledlabel{C_2} (manipulation of timestamp integrity) is detailed in the Appendix~\ref{sec:attack-c2}.

\subsubsection{PTP-based \attname Attack}
We instantiate the threat model of Section~\ref{sec:threat-model} with capability \circledlabel{C_1}: the attacker does not modify message payloads or timestamps directly but compromises the \emph{time-distribution plane}---the PTP-based clock synchronization used by all nodes. As shown in Fig.~\ref{fig:testbed_hardware}, we introduce a fourth, malicious node (Raspberry Pi) into the same Ethernet/PTP domain alongside the camera, LiDAR, and fusion nodes. The malicious node does not publish sensor data; it participates only in PTP. By exploiting the Best Master Clock Algorithm (BMCA), the attacker advertises a superior clock quality, is elected as Grandmaster (GM), and thus dictates the global time reference to which all legitimate nodes (and hence their local timestamps $t_S^{(i)}$) are bound. This takeover is protocol-compliant and transparent to the fusion stack.

\paragraph{Attack Mechanism}
After assuming the GM role, the attacker periodically forces the GM's advertised time to step \emph{backward} by a controlled amount $\delta_S$ (e.g., from $t_{\mathrm{attack}}$ to $t_{\mathrm{attack}} - \delta_S$), then lets time advance again naturally. In our ROS~2 pipeline, sensors stamp each message with their local wall clock at publish time; that value is exactly the \emph{local timestamp} $t_S^{(i)}$ used by the fusion node (Section~\ref{sec:ad-system-model}). A backward GM step propagates to slave nodes via PTP, so their system clocks---and thus the timestamps they attach---jump backward. Because ROS~2 disallows publishing messages with timestamps in the past relative to the last published time, sensor nodes effectively stop publishing for the real-time duration $\delta_S$ (a transient denial-of-service). When clocks later advance again and ``catch up'' to real time, sensors resume publishing with local timestamps that have moved forward. Critically, the \emph{universal capture times} $u_S^{(i)}$ of newly produced messages are now at least $\delta_S$ seconds ahead of the last pre-step message, yet the \emph{local timestamps} $t_S^{(i)}$ can lie in a narrow window that overlaps earlier timestamps. Thus, measurements from \emph{distinct physical moments} (separated in universal time by $\ge \delta_S$) are mapped into a \emph{narrow} local-timestamp window---\emph{temporal compression} on the reported timeline. The UCO $\phi_S^{(i)}$ becomes large, violating Assumption~\ref{assum:timestamp-fidelity}. Because camera and LiDAR nodes follow their own PTP servo and scheduler dynamics, they recover at different rates and instants; the aligner can therefore form pairs $(i,j)$ whose reported R-RTM $\tilde{\Delta}^{(i,j)}$ remains within the slop $\tau$, while the actual S-RTM $\Delta^{(i,j)}$ and the S-ATM $\alpha_C^{(i,j)}$, $\alpha_L^{(i,j)}$ are large. The attacker can tune $\delta_S$ and the step cadence to trade off transient DoS duration against the magnitude of induced misalignment and stealth.

\paragraph{Attack Impact}
We report results for a benign phase followed by an attack phase starting at $t = 5$\,s and lasting 5\,s, with the GM clock stepped backward by $\delta_S = 0.20$\,s every $0.22$\,s. Figure~\ref{fig:ptp-attack-c1-a} shows the effect on message transmission and on the local timestamps (header timestamps) reported by the camera and LiDAR nodes. Figure~\ref{fig:ptp-attack-c1-b} shows the effect on temporal alignment at the fusion node:  while the S-RTM and S-ATM values remained within the slop $\tau = 0.10$ under the benign period, their value spiked under the attack period, indicating that the aligner is emitting pairs that are semantically misaligned and/or stale. Fusing such pairs degrades perception; we quantify this in the following section. 

%% file: sections/s5_model_experiments.tex
\section{Evaluating \attname on MMF Models and End-to-End AD}
\label{sec:exp-setting-att}

This section evaluates the impact of \attname-induced temporal misalignment on state-of-the-art MMF-based perception models across KITTI and NuScenes, and further realizes \attname in an end-to-end AD simulator.

\subsection{Datasets}
\label{sec:datasets}
We evaluate our approach on two widely used multimodal AD datasets for 3D object detection and multi-object tracking:  

\vspace{3pt}
\noindent\textbf{KITTI Tracking Dataset.~} The KITTI Tracking Dataset~\cite{geiger2012we} was collected in Karlsruhe, Germany, across urban, suburban, and highway scenes. It provides a forward-facing RGB camera (1242$\times$375 resolution) and a Velodyne HDL-64E LiDAR operating at $\sim$10 Hz. The dataset contains 21 training and 29 test sequences with frame-level 3D bounding boxes and identity annotations for three main classes: cars, pedestrians, and cyclists.  

\vspace{3pt}
\noindent\textbf{NuScenes Dataset.~} The NuScenes Dataset~\cite{caesar2020nuscenes} was collected in Boston (USA) and Singapore, focusing on dense urban traffic under diverse conditions. It provides six surround-view RGB cameras, a Velodyne HDL-32E LiDAR (20 Hz), and five radars, with all annotations sampled at 2 Hz. The dataset consists of 1000 driving scenes, each 20 seconds long, with 3D bounding boxes and tracking IDs for 23 classes, including vehicles, pedestrians, bicycles, and traffic barriers. 

\begin{figure}[t]
    \centering
    \includegraphics[width=0.45\textwidth]{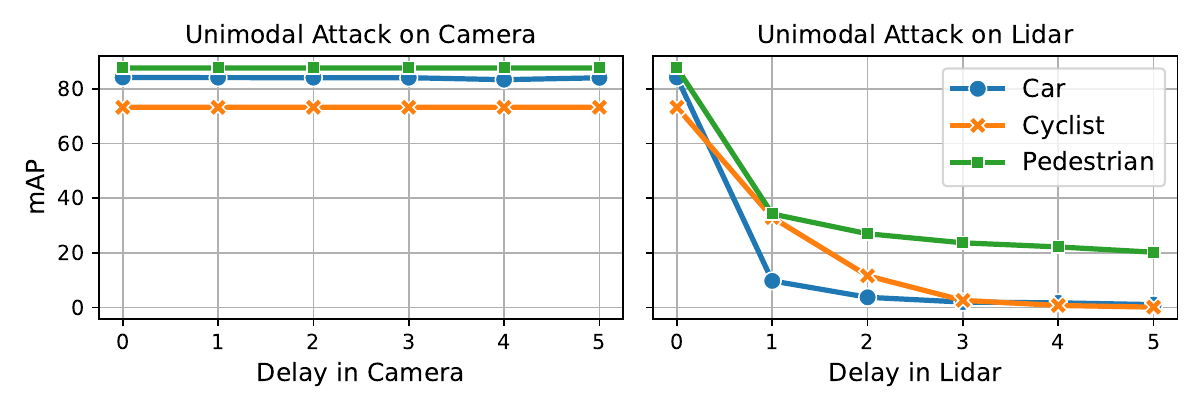}
    \caption{Uni-\attname attack impacts on 3D object detection performance of \textit{MVXNet} on KITTI dataset for different object classes.}    
    \label{fig:mvxnet_3d_detection_uni}
\end{figure}

\begin{figure}[t]
    \centering
    \includegraphics[width=0.5\textwidth]{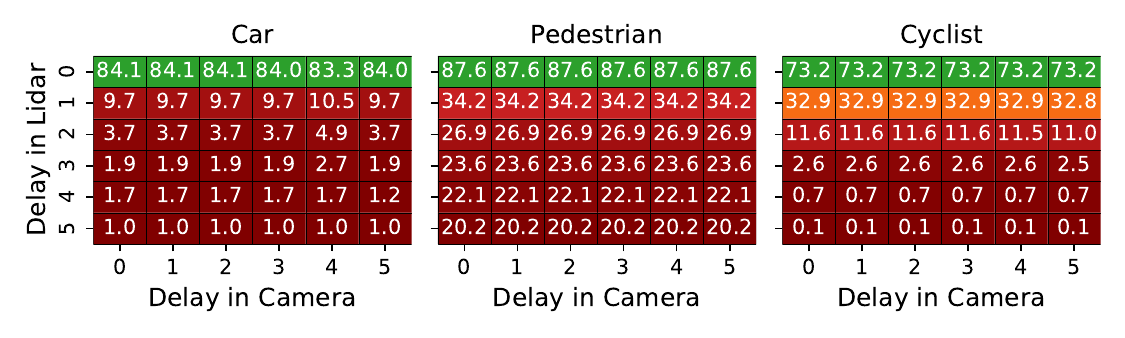}
    \caption{Mul-\attname attack impacts on 3D object detection performance of \textit{MVXNet} on KITTI dataset for different object classes.}
    \label{fig:mvxnet_3d_detection_multi}
\end{figure}

\begin{figure}[t]
    \centering
    \includegraphics[width=0.5\textwidth]{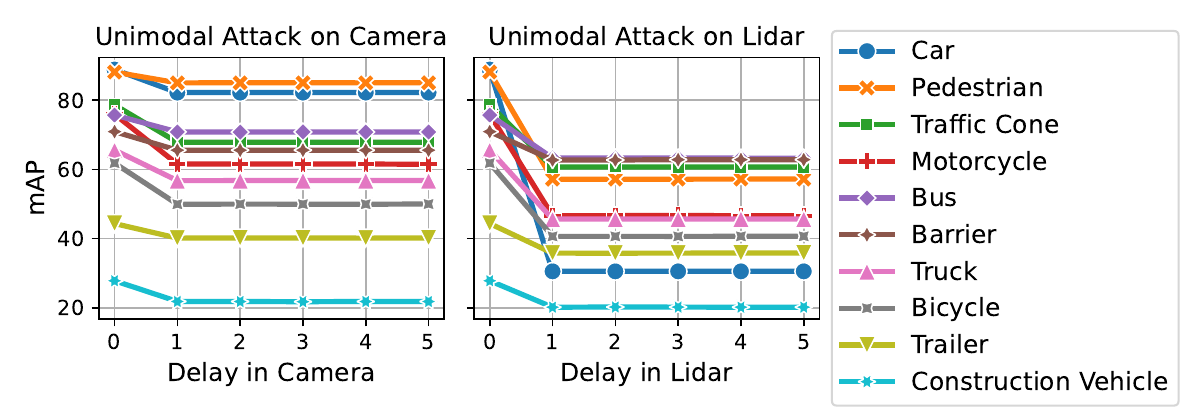}
    \caption{Uni-\attname attack impacts on 3D object detection performance of \textit{BEVFusion} on nuScenes dataset for different object classes.}
    \label{fig:bevfusion_3d_detection_uni}
\end{figure}

\subsection{Models}  
\label{sec:multimodal-models}  

To systematically evaluate the effect of \attname on MMF with different downstream tasks, we consider the following: 

\noindent\textbf{3D Object Detection:} We evaluate \attname on two representative MMF-based 3D object detection models. i) \textit{MVXNet}~\cite{sindagi2019mvx}, trained on the KITTI dataset,  is an early fusion-based architecture that projects LiDAR point clouds into pseudo-image space and fuses them with camera image features at the voxel level. ii) \textit{BEVFusion}~\cite{liu2023bevfusion} is a more advanced architecture, trained on the NuScenes dataset, that unifies multi-modal sensor inputs in the bird's eye view (BEV) representation space. BEVFusion is widely used in industry-grade AD software stacks, including Autoware. 

\noindent\textbf{Multi-Object Tracking (MOT):} For evaluating tracking performance under \attname, we consider \textit{MMF-JDT}~\cite{wang2024multi}, trained on the KITTI tracking dataset, is a joint detection and tracking model that incorporates early and mid-level fusion strategies to align image and point cloud features for improved object association over time. 

\begin{figure}[!t]
    \centering
    \includegraphics[width=0.495\textwidth]{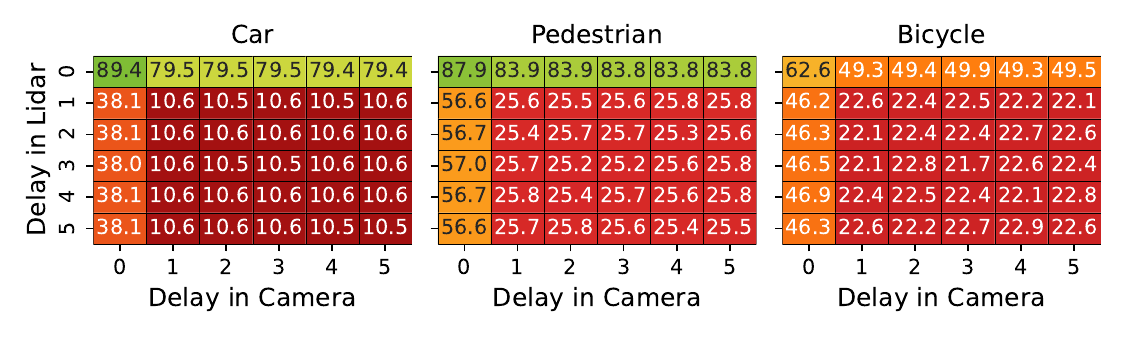}
    \vspace{-15pt}
    \caption{Mul-\attname attack impacts on 3D object detection performance of \textit{BEVFusion} on nuScenes dataset for three object classes.}
    \vspace{-10pt}
    \label{fig:bevfusion_3d_detection_multi_extended_three}
\end{figure}

\subsection{Attack Settings}
\label{sec:attack-settings}

To evaluate the impact of \attname attack on MMF models, we introduce controlled delays in one or both sensor modalities (camera and LiDAR), as mentioned in Table~\ref{tab:temporal_attacks}, and analyze the corresponding degradation in model performance. We systematically assess how different degrees and types of temporal misalignment affect the 3D object detection and multi-object tracking. 
Specifically, we use the S-ATM per modality as the delay parameter $\alpha \in \{0, 1, 2, 3, 4, 5\}$, where $\alpha=0$ represents a perfectly synchronized sensor, and $\alpha=5$ denotes a maximum delay of five frames for the affected modality. We consider both Uni-\attname and Mul-\attname attacks. In Uni-\attname attack, delay is introduced in either the camera or the LiDAR input while keeping the other modality synchronized. In Mul-\attname attacks, both camera and LiDAR streams are delayed independently, leading to varying degrees of temporal misalignment. We use the pretrained weights for the target model provided with the official implementations. We focus exclusively on attacking the test dataset by applying the defined temporal delays.


\subsection{Evaluation Metrics}  
To assess the impact of \attname, we analyze the model performance using task-specific evaluation metrics. For 3D object detection, we evaluate the models using mean average precision (mAP), which quantifies the accuracy of detected objects, as well as nuScenes Detection Score (NDS), particularly for the NuScenes dataset. For MOT, we use standard tracking metrics such as higher order tracking accuracy (HOTA), detection accuracy (DetA), association accuracy (AssA), multiple object tracking accuracy (MOTA), and Identity Switches (IDSW), which measure the effectiveness of object association.

\subsection{Software Implementation}
We implement and evaluate \attname using Python 3.8 and PyTorch, utilizing open-source frameworks including MMDetection3D~\citep{mmdet3d2020} and OpenPCDet~\cite{openpcdet2020}. Experiments were conducted on a server running Ubuntu 20.04.6 LTS with an Intel Xeon Gold 5520 (16 cores, 2.20GHz), 128GB RAM, and three NVIDIA RTX 6000 Ada GPUs.

\section{Evaluation Results}
\label{sec:eval-results}


This section presents the \attname attack impact on different MMF models and datasets, and discusses the key findings from the evaluation.

\subsection{Impact of \attname Attack on 3D Object Detection}
\label{sec:attname-3d-ods}

We investigate the effectiveness of the proposed \attname attack on multimodal 3D object detection using \textit{MVXNet} and \textit{BEVFusion}, evaluated on the KITTI and nuScenes datasets, respectively. We analyze the detection performance of 3D object detection across different object classes under varying levels of unimodal and multimodal sensor delay. Object detection models do not process sequential information; instead, their performance is affected only by frame-wise delays in each modality at any particular time, whether the delays are constant or random. Therefore, for simplicity and consistency, we only analyze the attack under the constant delay setting.


\begin{figure*}[!t]
\centering
\subfloat[Constant Delay Scenario]{%
    \includegraphics[width=0.99\textwidth]{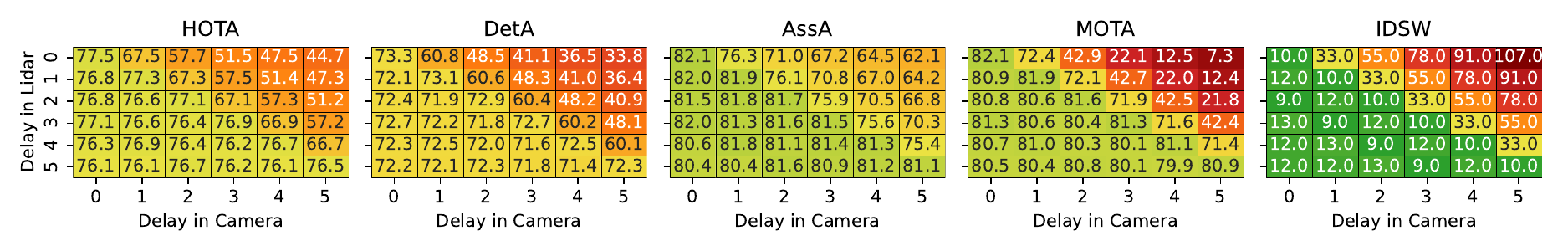}%
    \label{fig:mmf_jdt_constant_ext}
}
\hfil
\subfloat[Random Delay Scenario]{%
    \includegraphics[width=0.99\textwidth]{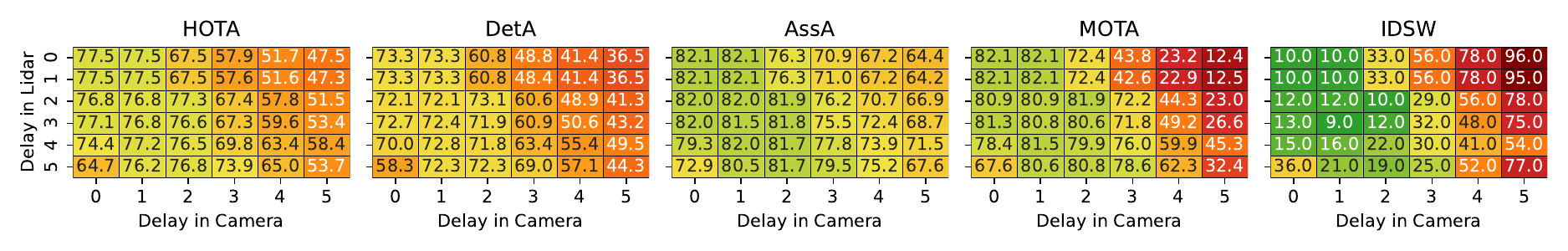}%
    \label{fig:mmf_jdt_random_ext}
}
\caption{\attname attack impacts on multi-object (car) tracking performance of \textit{MMF-JDT} on KITTI tracking dataset with respect to different metrics.}
\label{fig:mmfjdt_mot_multi_ext}
\end{figure*}

\subsubsection{MVXNet on KITTI Dataset}  
Fig.~\ref{fig:mvxnet_3d_detection_uni} presents the 3D object detection performance of \textit{MVXNet} under Uni-\attname attacks, where either the camera or LiDAR input is delayed independently. Under benign (zero-delay) conditions, \textit{MVXNet} achieves strong mAP across most object classes: approximately 84.1 for cars, 87.6 for pedestrians, and 73.2 for cyclists. The left plot shows that delaying the camera input alone--under Uni-\attname camera attacks---has minimal effect on performance across all classes, with nearly constant mAPs. In contrast, the right plot highlights the model's high sensitivity to LiDAR delays: a 1-frame delay causes the car mAP to collapse from 84.1 to 9.7 ($\downarrow$88.5\%), pedestrian mAP from 87.6 to 34.2 ($\downarrow$60.9\%), and cyclist mAP from 73.2 to 32.9 ($\downarrow$55.1\%), with further degradation as delay increases.  This indicates that LiDAR data is significantly more critical than camera input in \textit{MVXNet}’s perception pipeline; hence, \textit{MVXNet}'s high vulnerability against Uni-\attname attack against LiDAR. However, Fig.~\ref{fig:mvxnet_3d_detection_multi} shows the mAP heatmaps across combinations of camera and LiDAR delays under Mul-\attname attacks. Although the 1-frame LiDAR delay drops mAP from 55.1\% to 88.5\% for different objects, camera delay has almost no effect, indicating the dominance of LiDAR in \textit{MVXNet}.

\begin{keyfinding}
\textbf{Key Findings.~} \textit{MVXNet} heavily depends on LiDAR input for 3D object detection. Camera delay, under both Uni- or Mul-\attname attack, has almost no effect, but even minor LiDAR misalignment leads to severe performance degradation. When both modalities are delayed, the impact is still only dominated by the LiDAR stream. 
\end{keyfinding}

\subsubsection{BEVFusion on nuScenes Dataset}
Fig.~\ref{fig:bevfusion_3d_detection_uni} shows \textit{BEVFusion}’s 3D detection performance under Uni-\attname attacks. With no delay, the model achieves high mAP across most classes---approximately 88 for car and pedestrian, and slightly lower for the rest. With a 1-frame camera delay, mAP drops slightly across all classes (for instance, car mAP drops by 7.4\%), but remains stable with further delays, showing \textit{BEVFusion} is slightly vulnerable against Uni-\attname camera attacks. 
In contrast, introducing a 1-frame LiDAR delay results in a substantial reduction in mAP for specific object classes, with performance dropping by 65.6\% for cars (from 88.8 to 30.5), 35.1\% for pedestrians (from 88.1 to 57.2), and 38.6\% for motorcycles (from 75.9 to 46.6).
Although performance remains stable with further LiDAR delays, this finding shows \textit{BEVFusion}'s higher vulnerability against Uni-\attname LiDAR attacks. 
Similarly, Fig.~\ref{fig:bevfusion_3d_detection_multi_extended_three} illustrates \textit{BEVFusion}'s performance under Mul-\attname attacks for the three different objects (complete list of object is in Fig.~\ref{fig:bevfusion_3d_detection_multi_extended}). Although delaying the camera alone has a minimal impact on performance, combining it with a delay in the LiDAR stream results in a significant drop in mAP. 
For instance, a 1-frame delay in the camera or LiDAR stream reduces car mAP by 7.4\% and 65.5\%, respectively. However, when both modalities are delayed simultaneously by one frame, the mAP drops dramatically by 89.2\% (from 88.8 to 9.6), highlighting a substantial compounding effect. This trend is consistently observed across all object classes. 


\begin{keyfinding}
\textbf{Key Findings.~} BEVFusion is slightly affected by camera delays but is highly affected by LiDAR delays, further underscoring LiDAR's dominating role in 3D object detection. However, the impact becomes significant if both sensors are delayed simultaneously, even just by one frame.
\end{keyfinding}



\begin{figure}[!t]
\vspace{-10pt}
\centering
\subfloat{
    \includegraphics[width=0.495\textwidth, clip, trim=10 0 0 0]{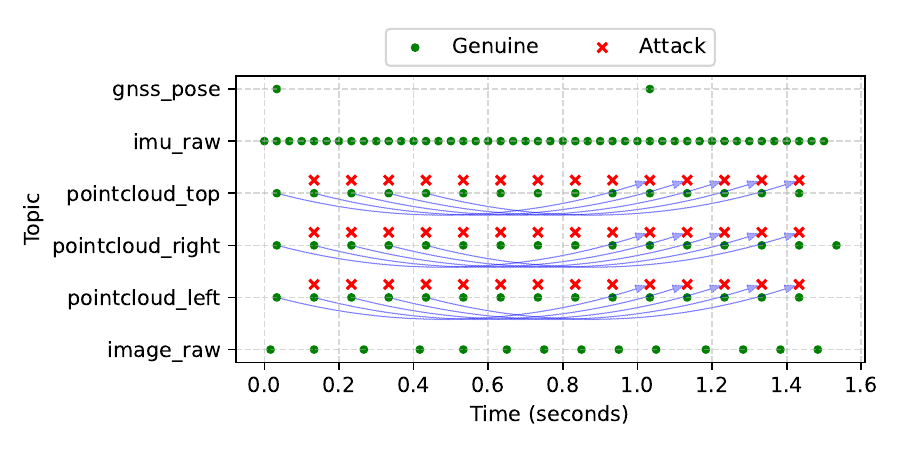}
    \label{fig:autoware-attack-recorded-data-a}
}
\caption{Demonstration of replay-style \attname attacks where malicious LiDAR messages closely follow genuine ones, increasing alignment likelihood with the malicious ones.}
\label{fig:autoware-attack-recorded-data}
\end{figure}

\begin{figure*}[t]
\centering
\subfloat[False negative scenario.]{
    \includegraphics[width=0.495\textwidth, clip, trim=0 10 0 0]{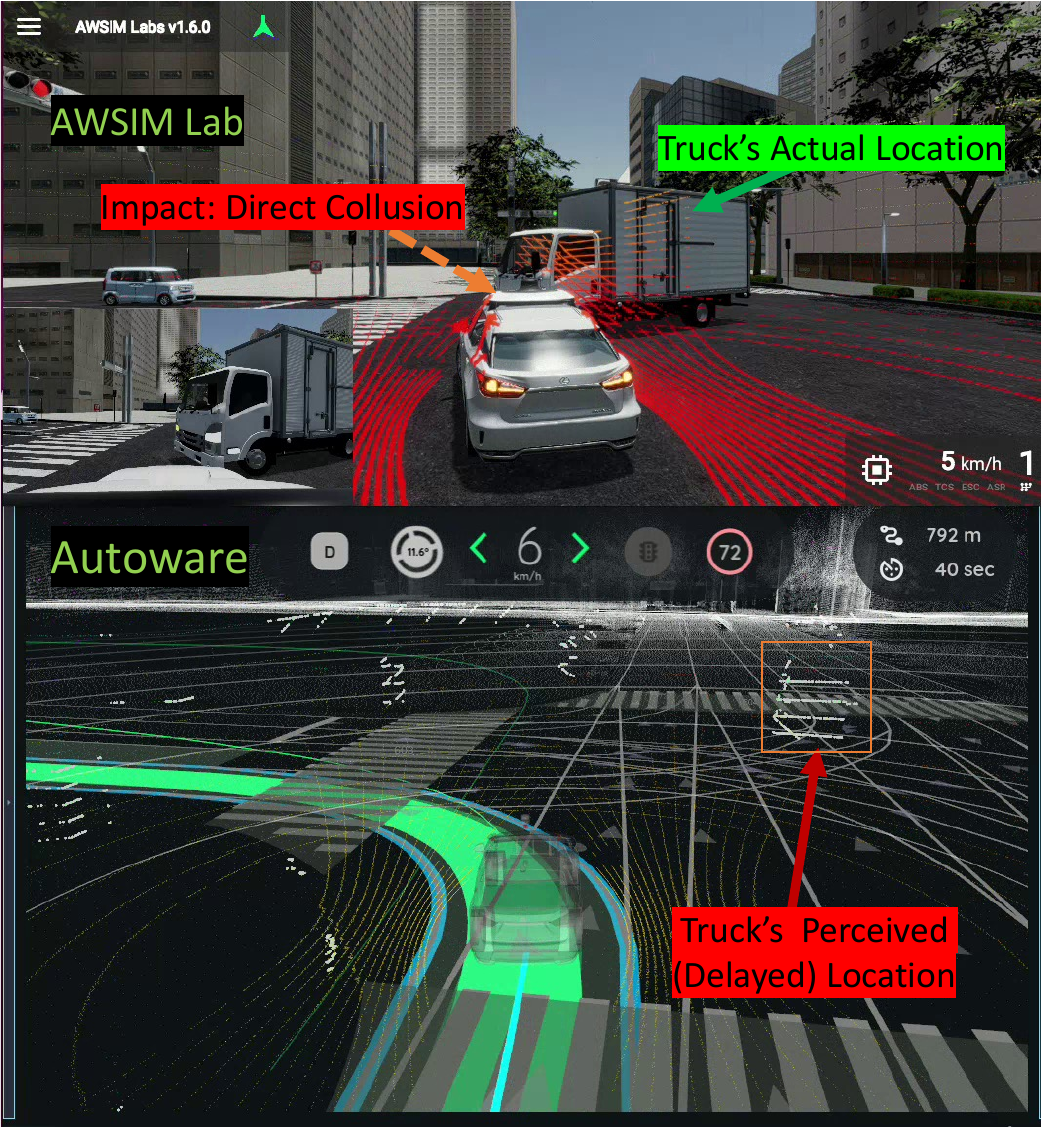}
    \label{fig:autoware_false_negative}
}
\subfloat[False positive scenario.]{
    \includegraphics[width=0.495\textwidth, clip, trim=0 10 0 0]{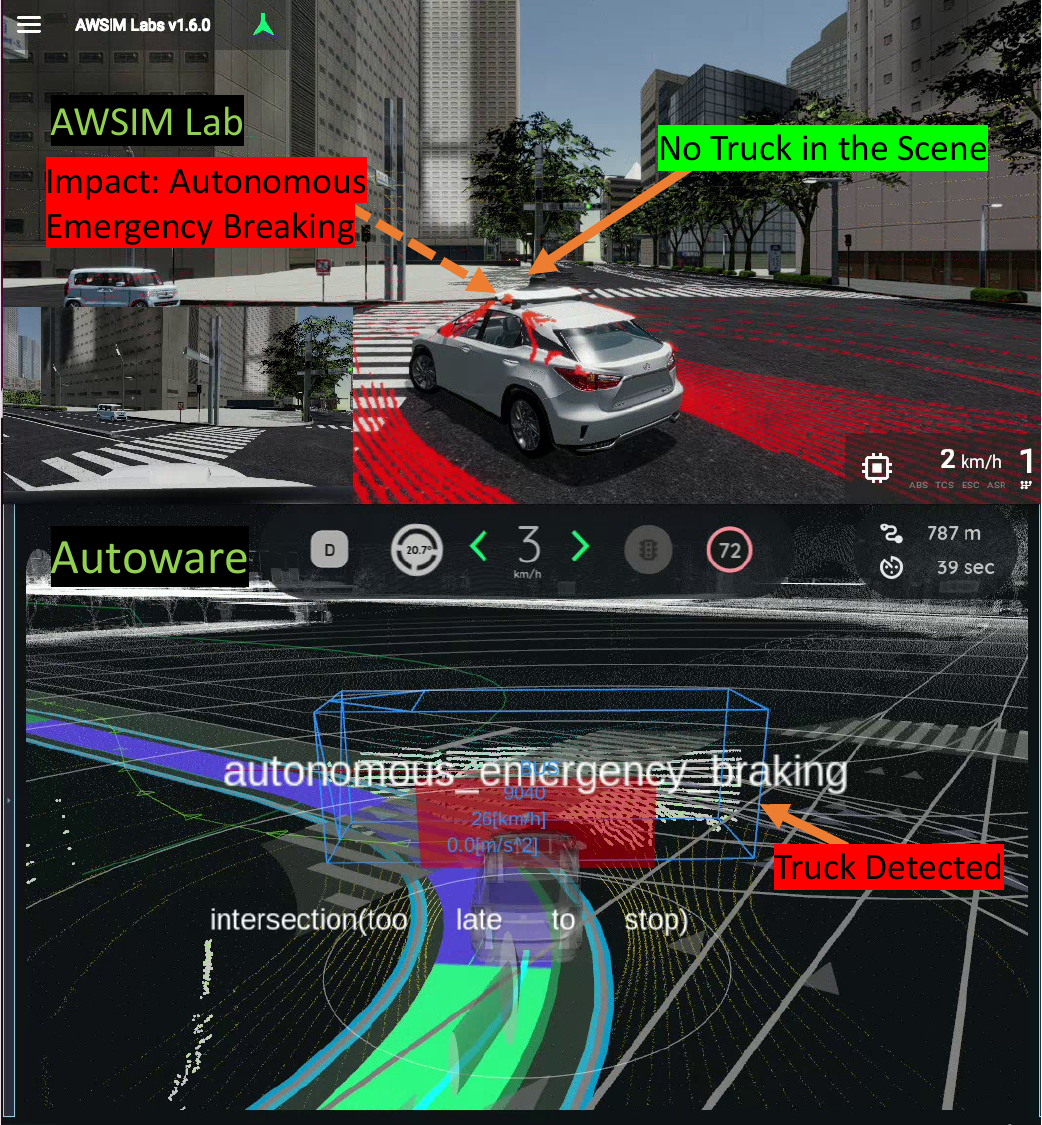}
    \label{fig:autoware_false_positive}
}
\caption{
Impact of delayed yet valid LiDAR data on Autoware. a) The ego misses the oncoming truck, causing a head-on collision, and b) the ego vehicle brakes for a non-existent object from delayed data.}
\label{fig:autoware-simulation}
\end{figure*}

\subsection{Impact of \attname Attack on Multi Object Tracking}
\label{sec:attname-mot}

We investigate the effectiveness of the proposed \attname attack on MOT algorithm using \textit{MMF-JDT} on the KITTI dataset. We analyze the tracking performance for cars under varying levels of delays under Uni- and Mul-\attname  attack scenarios.

\subsubsection{MMF-JDT on KITTI Dataset}  
This part studies the impact of \attname attacks on \textit{MMF-JDT} evaluated on the KITTI tracking dataset. For this evaluation, we consider both types of delays: constant (Fig.~\ref{fig:mmf_jdt_constant_ext}) and random (Fig.~\ref{fig:mmf_jdt_random_ext}). Across both attack scenarios, the tracking performance declines as camera delay increases. 
Along with other tracking metrics, MOTA and IDSW suffer noticeable degradation as camera delays increase under Uni-\attname attacks. For instance,  IDSW---a metric that captures identity switches and ideally should be low, increases dramatically under increasing camera delays, underscoring the disruption in tracking consistency caused by \attname attacks.


Furthermore, the values of different metrics across the heatmaps of Fig.~\ref{fig:mmf_jdt_constant_ext} suggest that while camera delay under Uni-\attname deteriorates the performance, delaying both the camera and LiDAR by exactly the same delay (\textit{i.e.,} constant delay scenario) under Mul-\attname attack diminishes the attack impact and mostly retains the performance. 
On the other hand, heatmaps in Fig.~\ref{fig:mmf_jdt_random_ext} show that Mul-\attname attacks with random delays remain effective as different delays in both modalities break the sequence, making object tracking considerably more difficult. For instance, under the Mul-\attname constant attack scenario with a five-frame delay, MOTA decreases by only 1.4\%, whereas the random attack scenario results in a 60.5\% drop.


\begin{keyfinding} 
\textbf{Key Findings.~} In contrast to 3D object detection tasks, MOT appears to rely more heavily on camera inputs. This may be attributed to the fact that MOT does not require precise 3D bounding boxes; instead, the rich texture information in camera images may offer more effective contrastive representations than sparse point clouds. As a result, \attname attacks can substantially impair MOT performance, particularly under Uni-\attname (camera delay) and Mul-\attname with random delay scenarios. 
\end{keyfinding}

\subsection{Simulation-Based End-to-End AD Setup}
We conduct our experiments using {Autoware}, an open-source full-stack autonomous driving framework. Autoware is widely adopted in both commercial and public-sector deployments, including {Level 4 autonomy trials} and government-funded programs (e.g., the U.S. Department of Transportation's CARMA\textsuperscript{SM} initiative). Its extensive use in real-world systems makes it a realistic and representative platform for evaluating the safety and robustness of autonomous driving pipelines.
To simulate real-world driving scenarios in a reproducible and controllable setting, we integrate Autoware with {AWSIM Lab}---a Unity-based, open-source simulator developed as part of the Autoware ecosystem---that provides high-fidelity urban environments and realistic sensor simulation. The simulated vehicle is equipped with a representative sensor suite including \textit{GNSS}, \textit{IMU}, three \textit{Velodyne VLP-16 LiDARs}, and a \textit{traffic light camera}. 
ROS~2 (Humble) works as the middleware, enabling seamless and real-time communication between AWSIM's simulated sensors and Autoware full autonomy stack, allowing us to evaluate the impact of \attname attack in a safe yet realistic environment. Our experiments are conducted on {Tokyo’s Nishishinjuku district} road map.

\textbf{\attname Attack Impact on AD Simulation.} We demonstrate the \attname attack in a full-stack autonomous driving pipeline, targeting the LiDAR sensor under attacker capability \circledlabel{C_3} (see Section~\ref{sec:threat-model}). Given LiDAR’s dominant role in 3D perception, the attacker impersonates three legitimate LiDAR nodes by subscribing to their respective ROS~2 topics and monitoring inter-frame intervals (10 Hz) to predict the next transmission times. The attacker stores recent point cloud messages and, at the time of attack, publishes forged messages with previously captured data with updated timestamps---just before the expected legitimate message. This increases the likelihood that the forged message is selected by the time-based synchronizer if it aligns more closely with the timestamps of other modalities (e.g., IMU, camera). Figure~\ref{fig:autoware-attack-recorded-data} illustrates how these delayed messages are positioned and transmitted on the same topics, effectively impersonating genuine LiDAR messages in real time.


When the forged messages are utilized in the downstream tasks, in the most severe case, the system completely misses the presence of an actual oncoming vehicle, resulting in a head-on collision at an intersection—despite nearby objects being within sensor range (Fig.~\ref{fig:autoware_false_negative}). This constitutes a false negative perception failure with life-threatening implications.
In another scenario, delayed LiDAR data causes the ego vehicle to perceive a non-existent obstacle---a vehicle that has already passed---leading to unnecessary emergency braking (Fig.~\ref{fig:autoware_false_positive}). This false positive event can create rear-end collision risks. In both cases, the outdated sensor data was used to repeatedly overwrite fresh messages, degrading the temporal integrity of the perception pipeline. 


Beyond object-level failures, we observe broader impacts across the autonomy stack. The tracker may assign separate IDs to genuine and delayed instances of the same object, interpreting them as distinct entities. Similarly, temporal inconsistencies introduced in the LiDAR stream desynchronize SLAM modules, leading to localization drift and control failures such as veering off-lane or collisions with curbs and roadside objects (as shown in Fig.~\ref{fig:slam_failure} in the Appendix).

%% file: sections/s7_discussion.tex
\section{Discussion}
\label{sec:discussion}

\subsection{Attack Limitations}

While the \attname attack demonstrates the vulnerability of multimodal perception systems to temporal misalignment, there are several limitations that constrain its applicability in real-world settings. First, the attack has not been validated on a physical vehicle platform, where additional practical challenges such as sensor noise, actuator delays, and system integration issues may affect both the feasibility and effectiveness of the attack.  
Besides, the attack assumes a high level of knowledge about the target system, including the sensors and network architecture. In practice, the attack requires adversarial access to the in-vehicle network. Although not trivial, prior work shows that physical access through the OBD-II port, remote exploitation of infotainment systems, or supply-chain insertion can provide such capabilities \cite{checkoway2011comprehensive}.  However, the adversary would still need to perform an exploration phase to gather this information, which introduces additional complexity and may limit the ability to execute the attack stealthily.  


Moreover, the evaluation does not account for real-world network-induced delays and variability, which could impact the targeted timing manipulation strategies. In actual deployment, the presence of stochastic latency and jitter may reduce the precision with which an attacker can control temporal misalignment, potentially diminishing the attack's effectiveness.  
Finally, the evaluation was conducted exclusively on an offline dataset with relatively low frame rates. Real-time systems typically operate at higher frame rates, and the metrics and thresholds used in this offline evaluation may not directly translate to real-world performance. Consequently, the practical impact of the attack on deployed autonomous systems may differ from the results observed in the experimental study.

\begin{figure}[t]
\centering
\subfloat{
    \includegraphics[width=0.24\textwidth, clip, trim=0 10 0 0]{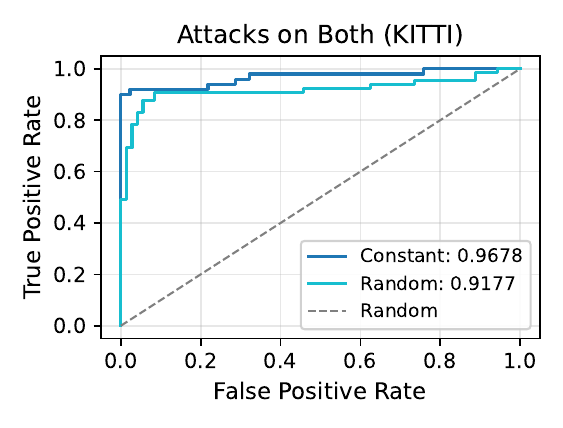}
    \label{fig:kitti}
}
\hspace{-15pt}
\subfloat{
    \includegraphics[width=0.24\textwidth, clip, trim=0 10 0 0]{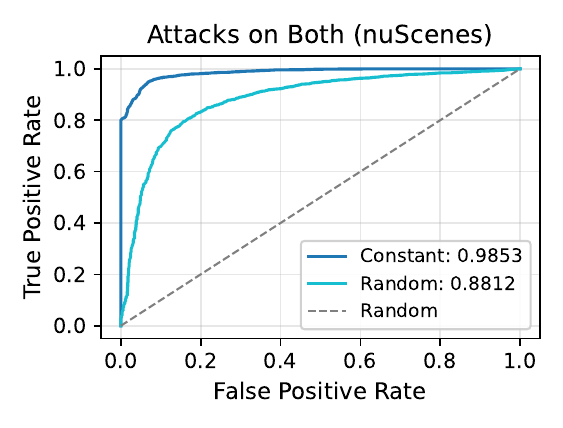}
    \label{fig:nus}
}
\caption{{\color{myblue}ROC performance of the OC-SVM–based temporal misalignment detector under mul-\attname attacks on KITTI (left) and nuScenes (right). Detection uses $3\times3$ cross-modal similarity matrices constructed from historic shared camera–LiDAR embeddings.}}
\label{fig:defense}
\end{figure}

\subsection{Defense Strategies Against \attname}

Hardening defenses can reduce the attack surface by securing sensor timestamps before fusion. This can be achieved via authenticated, hardware-anchored timestamps with cryptographic signatures, monotonic sequence numbers, and hardware-level clocks, such as network interface controller/SoC real-time clock (RTC). Combining multiple time sources---PTP, GNSS, and local RTCs---further strengthens temporal integrity. While these measures add overhead, they significantly raise the barrier to adversarial manipulation.

Detection defenses can identify temporal misalignments in real time. Techniques include intermodality temporal-consistency analysis of embeddings, kinematic cross-checks using IMU, odometry, and other controller area networks (CAN) signals, and statistical monitoring of monotonicity, jitter, and freshness counters, before fusion. These mechanisms can detect out-of-order or replayed frames, allowing rapid response to potential attacks.

Mitigation techniques can limit the impact of detected misalignments on vehicle control. Delay-aware adaptive fusion compensates for inter-modal lags, weighs sensor inputs, and computes confidence scores. If confidence is low, conservative measures, such as slowing down, increasing spacing, or lowering autonomy, can be applied, ensuring safety even under temporal attacks.

{\color{myblue}\subsubsection{One-Class SVM Detection Defense}

To detect \attname attack, we design an unsupervised defense based on cross-modal similarity modeling and One-Class SVM (OC-SVM) classification. We first learn a \emph{shared representation} that projects camera and LiDAR features into a common embedding space. For each of the fusion pairs, we compute the pairwise cross-modal similarity and form a temporal similarity matrix over a sliding window of $w$ consecutive messages ($w=3$). These $w \times w$ similarity snapshots are flattened and used as inputs to the OC-SVM. To prevent data leakage, the detector is trained exclusively on the first half of benign (attack-free) test data and evaluated on the remaining half, which contains injected attacks.

Figure~\ref{fig:defense} reports the ROC curves of the OC-SVM detector under mul-\attname attacks (as conducted on Section~\ref{sec:hil}) on KITTI and nuScenes. Under the \emph{constant attack}, the detector achieves strong performance on both datasets (AUC $=0.9678$ on KITTI and $0.9853$ on nuScenes), as the prolonged freezing of timestamps introduces persistent degradation in cross-modal similarity that deviates from the learned benign manifold. In contrast, detection under the \emph{random attack} is noticeably weaker (AUC $=0.9177$ on KITTI and $0.8812$ on nuScenes). This gap arises because random forward and backward clock perturbations intermittently preserve short-term alignment structure, allowing portions of the similarity matrices to remain consistent with benign temporal patterns.

These results expose fundamental limitations of learning-based detectors under temporally irregular attacks. While constant offsets induce persistent distortions that are easier to detect, random offsets produce transient, stochastic misalignments that are difficult to distinguish from benign timing jitter. More broadly, detectors trained on limited temporal patterns risk bias toward the training distribution and may fail to generalize to unseen driving contexts and adaptive attacks. This motivates the need for more robust and generalized defenses, as well as the integration of additional modalities such as CAN bus and IMU data to improve sensing fidelity and strengthen cross-modal integrity under realistic and adversarial conditions.
}

%% file: sections/s8_conclusion.tex
\section{Conclusion}
\label{sec:conclusion}

This work presents \attname, a temporal misalignment attack that exploits synchronization vulnerabilities in multimodal perception systems for autonomous driving. Through extensive evaluations on state-of-the-art 3D object detection and multi-object tracking models, we uncover modality-specific vulnerabilities: 3D detection models are predominantly reliant on LiDAR and suffer severe degradation---with up to 88.5\% drop in mAP---from even a single-frame LiDAR delay, while MOT model exhibits heightened sensitivity to camera stream disruptions, with MOTA dropping by 73\% under just three-frame camera delays. 
These findings highlight the critical need for synchronization-aware design in perception architectures and emphasize the importance of robust temporal consistency checks in safety-critical autonomous systems.


\section*{ACKNOWLEDGMENT}
This work was supported in part by the Office of Naval Research under grants N00014-24-1-2730, and the National Science Foundation under grants 2235232, 2312447, 2247560, 2154929, 2154930, 2238635, 2403758, 2509636, 2312794,
and a fellowship from the Amazon-Virginia Tech Initiative for Efficient and Robust Machine Learning.

\section*{LLM Usage Disclosure.} LLMs were used for editorial purposes in this manuscript, and all outputs were inspected by the authors to ensure accuracy and originality.

%% file: sections/s9_appendix.tex
\appendices
\section{Extended Results on BEVFusion Model}

\begin{figure*}[!t]
    \centering
    \includegraphics[width=0.975\textwidth]{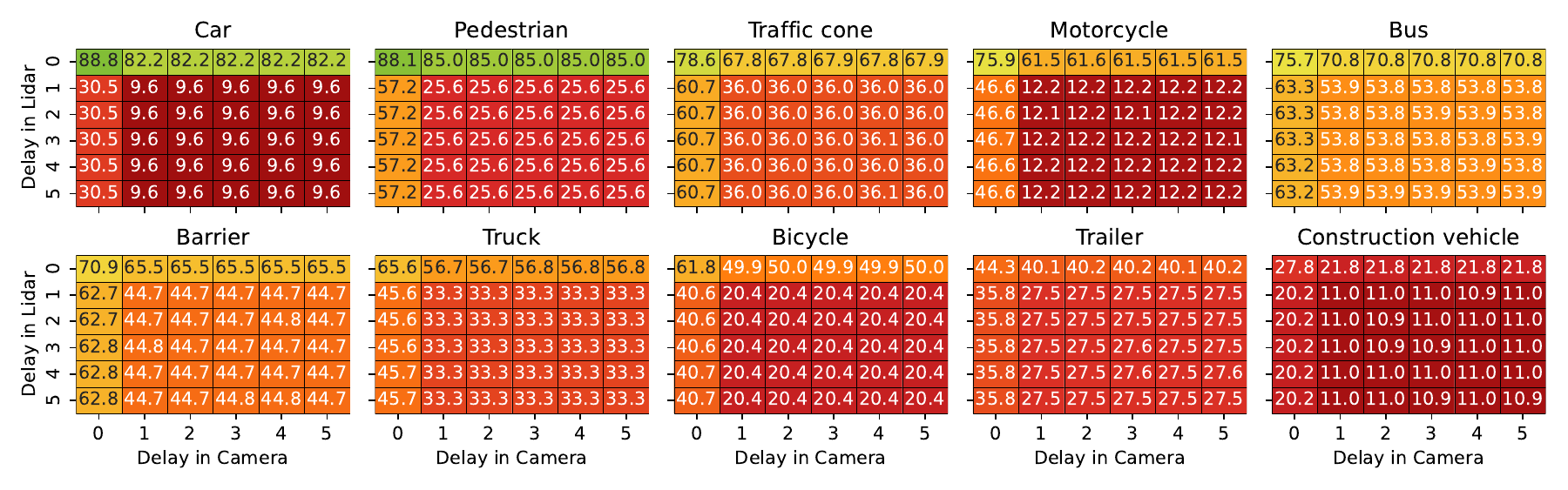}
    \caption{Mul-\attname attack impacts on 3D object detection performance of \textit{BEVFusion} on nuScenes dataset for all different object classes.}
    \label{fig:bevfusion_3d_detection_multi_extended}
\end{figure*}

\begin{figure}[!t]
    \centering
    \includegraphics[width=0.5\textwidth]{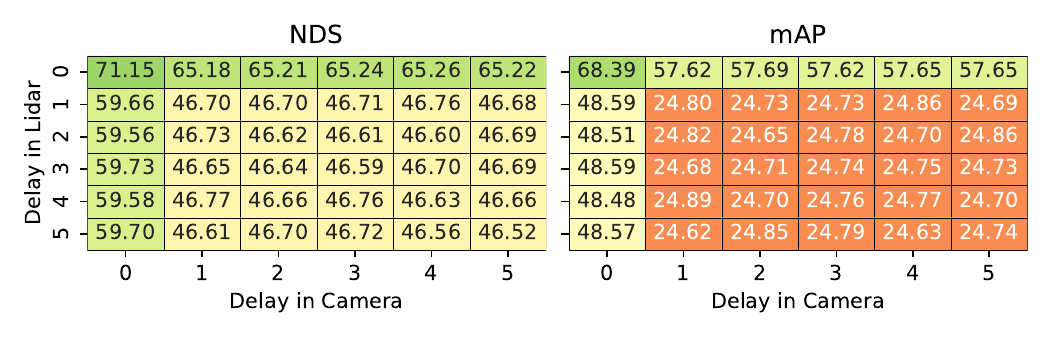}
    \caption{Overall Mul-\attname attack impacts on 3D object detection performance of \textit{BEVFusion} on nuScenes dataset for all the object classes regarding i) NDS, ii) mAP.}
    \label{fig:bevfusion_3d_detection_nds_extended}
\end{figure}

Fig.~\ref{fig:bevfusion_3d_detection_multi_extended} shows the Mul-\attname attack impacts on 3D object detection performance of \textit{BEVFusion} on the nuScenes dataset for all different object classes. Figure~\ref{fig:bevfusion_3d_detection_nds_extended} further shows the impact of Mul-\attname on the performance of \textit{BEVFusion} on the nuScenes dataset regarding average mAP and NDS score. 

\section{\attname Attack to Manipulate Timestamp Integrity.}
\label{sec:attack-c2}

\begin{figure}[t]
    \centering
    \includegraphics[width=0.495\textwidth]{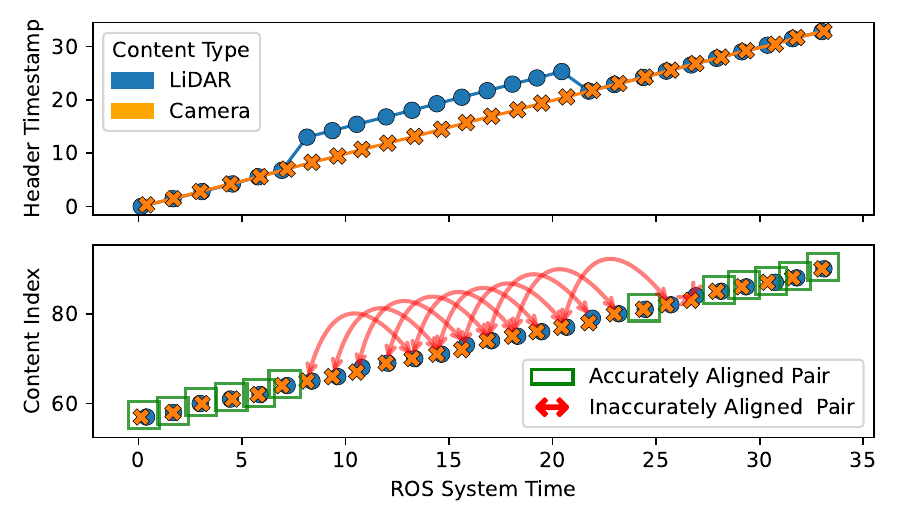}%
    \caption{ROS System time vs (top) Timestamp of Camera and LiDAR messages, and (bottom) Message content of Camera and LiDAR messages with }
    \label{fig:fusion_pairs}
\end{figure}

In this experiment, we assume that the attacker is targeting the LiDAR sensor with the attacker capability \circledlabel{C_2} (see Section~\ref{sec:threat-model}). As shown in Fig. \ref{fig:fusion_pairs}(top), the attacker first publishes six LiDAR messages with benign timestamps; hence, they are very close to the camera messages' timestamps. The attacker then deliberately introduces a constant offset of 5 seconds to the timestamps of the next eleven LiDAR messages, while continuing to transmit genuine, real-time data. As a result, the timestamps of these LiDAR messages are abruptly increased, as depicted in the figure. 

\begin{figure*}[t]
\centering
\includegraphics[width=0.99\textwidth, trim=0 100 100 325, clip]{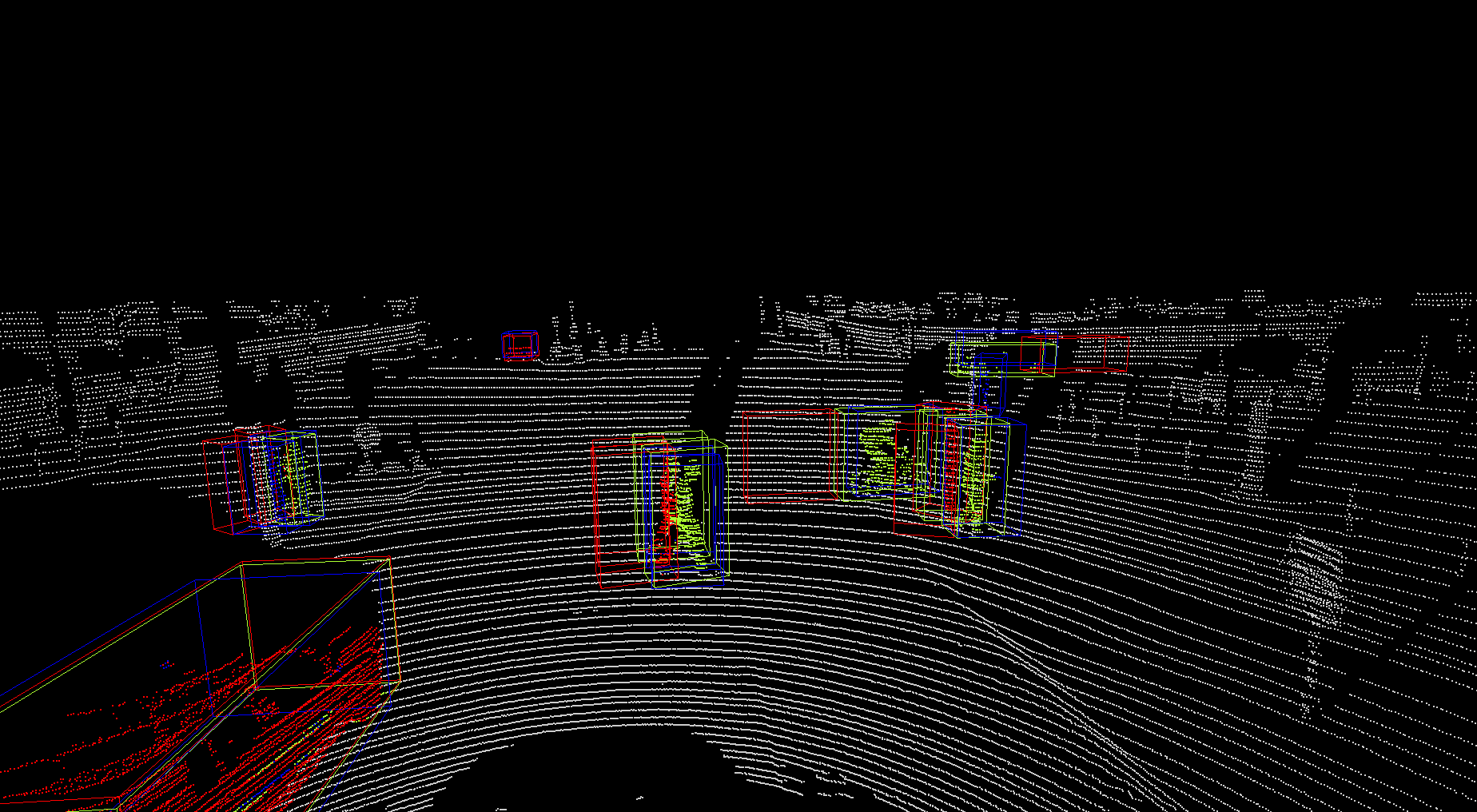}
    \label{fig:attack-lidar}

\caption{
Visualization of the impact of the \attname attack on a camera--LiDAR fusion-based perception model.
Ground-truth boxes are shown in \textcolor{blue}{blue}, benign predictions in \textcolor{green!60!black}{green}, and predictions under \attname in \textcolor{red}{red}.
MMF-based fused predictions are overlaid on the current LiDAR frame.
While benign predictions closely track the ground truth, delaying the LiDAR stream by five frames under \attname induces cross-sensor temporal misalignment, shifting predicted bounding boxes toward objects’ past locations and causing some objects to be missed.
This behavior suggests an overreliance on LiDAR inputs relative to the camera stream.
A corresponding Camera-plane visualization is provided in the Appendix (Fig.~\ref{fig:example_dejavu_attack_v2}).
}

\label{fig:example_dejavu_attack_v3}
\end{figure*}

In the bottom panel of Fig.~\ref{fig:fusion_pairs}, the message content is accurately synchronized for both camera and LiDAR pairs, for all benign timestamps, which are indicated by green rectangles. When the attacker sends manipulated timestamps, the synchronizer forces camera messages to align with LiDAR messages inaccurately, which are shown using red arrows in the figure. This demonstrates that, although the contents of camera and LiDAR messages were sequential, the synchronizer fails to align them accurately during the attack, as it prioritizes the timestamps of the messages. The impact of this attack is similar to the attack we see in Fig.~\ref{fig:example_dejavu_attack_v2} and Fig.~\ref{fig:example_dejavu_attack_v3}.

\section{Impact of \attname on SLAM}

\begin{figure*}[h]
    \centering
    \includegraphics[width=0.95\linewidth, clip, trim=0 100 0 100]{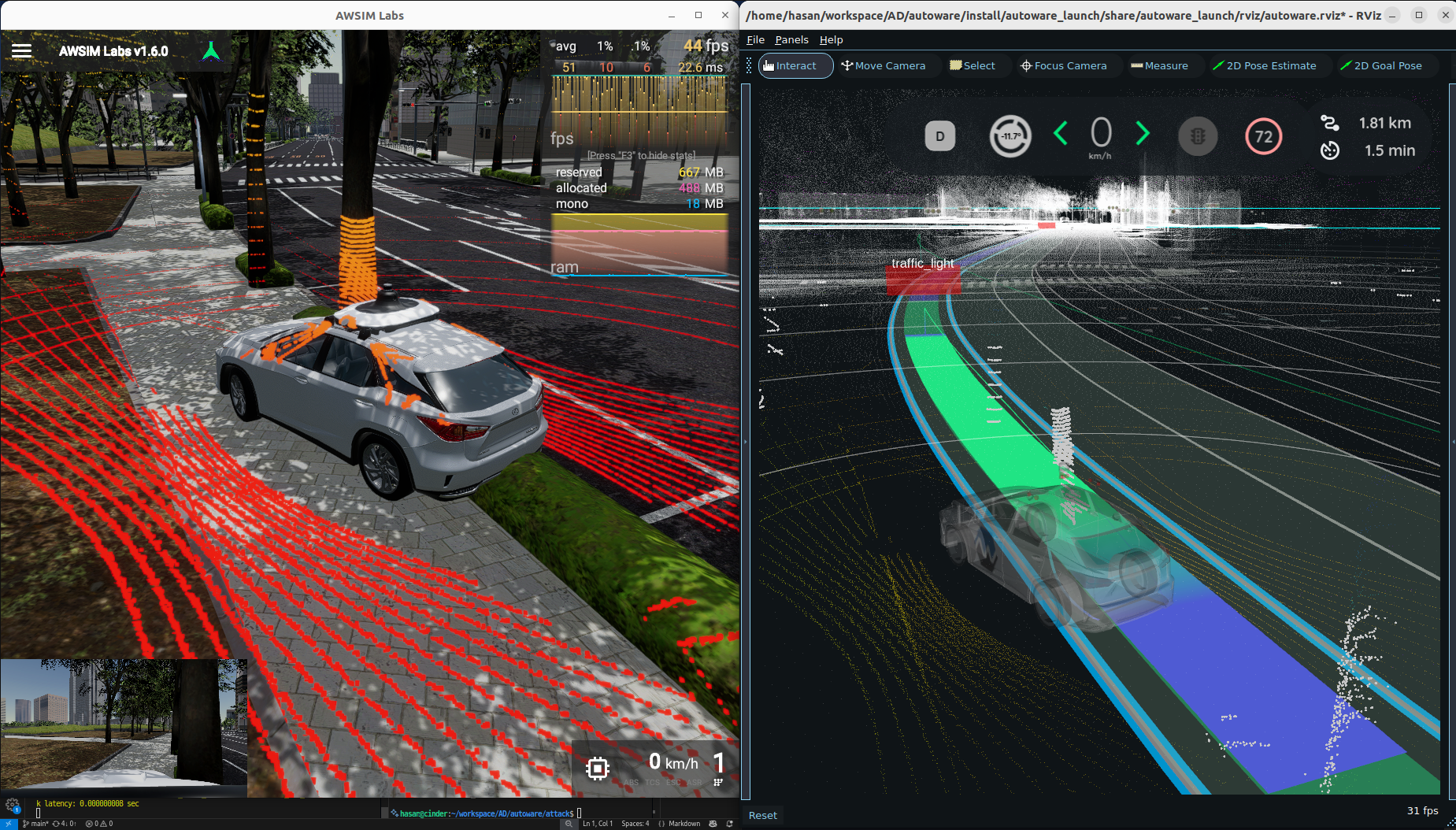}
    \caption{Impact of delayed but valid LiDAR data on Autoware: the induced SLAM drift propagates to the planning module, ultimately causing the vehicle to collide with the curb.}
    \label{fig:slam_failure}
\end{figure*}

Fig.~\ref{fig:slam_failure} illustrates the impact of \attname attack on SLAM due to the delay in the LiDAR stream. 

{\color{myblue}
\section{Hardware-in-the-Loop Testbed Description}

\subsection{Hardware Architecture}

Our testbed emulates an in-vehicle automotive Ethernet network using three Raspberry Pi 4B units, each equipped with Time-Sensitive Networking (TSN) Network Interface Cards (NICs), connected through RAD-Moon media converters to a RAD-Jupiter TSN-capable switch. This architecture forms a realistic automotive Ethernet backbone where sensor data streams traverse the same network path as they would in a production vehicle, enabling controlled studies of temporal misalignment attacks and their impact on multimodal fusion.

The three Raspberry Pi units serve distinct roles in the data pipeline. Two source nodes replay prerecorded camera and LiDAR datasets, publishing them as ROS~2 \texttt{sensor\_msgs/Image} and \texttt{sensor\_msgs/PointCloud2} messages, respectively. The third unit acts as an ECU fusion node, subscribing to both sensor streams, performing temporal synchronization, and logging timing metrics for post-processing analysis. Each Raspberry Pi is connected to a TSN-capable Ethernet NIC that provides hardware timestamping and 100BASE-T1/TSN functionality, bridging conventional Ethernet interfaces with the automotive Ethernet domain. A fourth node is used to take part in PTP communication and to advertise a superior clock quality, and eventually, to be selected as Grandmaster (GM). 

Three RAD-Moon media converters serve as transparent Layer~1/Layer~2 bridges, converting between 100BASE-T1 automotive Ethernet and conventional 10/100BASE-TX Ethernet. Two converters connect the source nodes to the network backbone, while a third converter bridges the ECU node. Each converter operates in full-duplex mode, automatically forwarding traffic between its two PHY interfaces while preserving link timing characteristics. The RAD-Jupiter switch, based on the Marvell 88Q5050 ASIC, serves as the central aggregation point with multiple 100BASE-T1 ports and an integrated AVB/TSN stack, routing time-sensitive traffic with deterministic latency guarantees. The physical connectivity follows a star topology centered on the switch, with camera and LiDAR data streams converging at the switch before being routed to the fusion ECU.

\subsection{Software Stack}

All Raspberry Pi units in the MMF pipeline run Raspberry Pi OS (64-bit) with vendor-provided TSN NIC drivers that enable correct PHY mode selection, hardware timestamping, and kernel-level optimizations for deterministic network behavior. The software stack uses ROS~2 (rclpy) for inter-node communication, employing standard message types (\texttt{sensor\_msgs/Image} for camera frames and \texttt{sensor\_msgs/PointCloud2} for LiDAR point clouds) with timestamps embedded in message headers via ROS~2's clock abstraction. Network configuration uses static IP addressing for deterministic routing, while the media converters and switch operate transparently at Layer~2, requiring minimal configuration beyond PHY mode selection.

\subsection{ROS~2 Node Architecture}

\subsubsection{Sensor Replay Nodes}

The camera and LiDAR nodes abstract prerecorded datasets as live sensor streams, enabling reproducible experiments with controlled timing characteristics. Both nodes employ a deterministic, time-based indexing mechanism that maps wall-clock time to frame indices through a shared seed function, ensuring synchronized frame selection across sensors while decoupling frame selection from ROS timer callbacks. This design makes frame ID selection a function of absolute time rather than message count, providing temporal consistency across experimental runs.

The camera node publishes images at a configurable rate (typically 10 Hz), embedding the current ROS~2 timestamp in each message header and encoding the frame index in the \texttt{frame\_id} field. The LiDAR node follows an identical pattern, publishing point cloud messages with synchronized timing. Additionally, the LiDAR node includes built-in support for temporal offset injection, allowing controlled studies of timestamp manipulation effects. This capability enables systematic injection of temporal misalignment without modifying the fusion node, supporting modular attack experimentation.

\subsubsection{Fusion and Synchronization Node}

The fusion node implements the ECU fusion logic, subscribing to both camera and LiDAR streams and aligning them temporally using ROS~2's \texttt{ApproximateTimeSynchronizer}. The synchronizer employs configurable queue buffering and temporal slop tolerance (typically $\pm 0.1$ second) to match messages with similar timestamps, handling out-of-order arrivals caused by network jitter or temporal attacks. Separate raw subscriptions are maintained for per-message logging, recording all received messages independently of the fusion timing logic to capture complete arrival statistics.

When the synchronizer identifies a matching camera-LiDAR pair within the temporal tolerance window, the fusion callback is triggered with both messages, which involves optional MMF-based perception for visualization. In that case, the synchronized image and point cloud data are then passed to the MVXNet perception model, which performs multimodal fusion and 3D object detection, enabling evaluation of how temporal misalignment affects fusion quality and detection accuracy.




\section{End-to-End Autoware Testbed Description}

\subsection{Simulation and System Configuration}
Our end-to-end evaluation is conducted using the Unity-based AWSIM Lab simulator tightly integrated with the open-source Autoware autonomous driving stack. AWSIM provides a high-fidelity digital twin of the ego vehicle, featuring photorealistic three-dimensional urban environments, realistic vehicle dynamics, and configurable automotive-grade sensors, including LiDAR, cameras, IMUs, and GNSS. Sensor extrinsics and vehicle geometry are specified through a URDF model, while a unified simulation clock ensures synchronized multi-sensor data generation. Native ROS~2 integration enables closed-loop operation by publishing simulated sensor streams and vehicle states and subscribing to drive-by-wire control commands from Autoware, thereby coupling perception, planning, and control in real time.

We employ a high-definition urban map of Tokyo’s Nishishinjuku district, which combines precise lane-level geometry, semantic Lanelet2 annotations, and stochastic traffic agents. The environment includes static assets (roads, buildings, and infrastructure) as well as dynamic actors (vehicles and pedestrians), enabling systematic evaluation across diverse traffic densities, intersection layouts, and environmental conditions. Semantic map annotations enforce traffic rules and directly support localization, prediction, and behavior planning, yielding a realistic urban driving workload.

\subsection{Perception}
Perception is primarily driven by LiDAR-based deep object detection. Raw point clouds are processed by a 3D neural network detector based on the CenterPoint/PointPillars family of architectures, which performs bird’s-eye-view object detection and semantic classification of surrounding traffic participants, including vehicles, pedestrians, and cyclists.

The detected objects are temporally associated and filtered by a multi-object tracking module based on Kalman filtering and probabilistic data association, yielding stable object identities and refined kinematic state estimates. In parallel, LiDAR returns are used to construct a dynamic occupancy grid that encodes free space and obstacles for downstream collision checking and trajectory optimization. Traffic-signal handling is performed through map-based stop-line association. Overall, this perception pipeline reflects a production-style LiDAR-centric sensing architecture widely used in modern autonomous driving systems.

\subsection{Planning}
The planning subsystem follows a hierarchical decision-making architecture. A mission planner computes a global route over the lane-level map using semantic navigation constraints. A behavior planner subsequently enforces traffic rules and selects high-level maneuvers such as lane following, lane changes, yielding, and stopping based on the current traffic context.

A motion planner then optimizes a dynamically feasible trajectory and velocity profile that minimizes tracking error and collision risk while satisfying kinematic and dynamic vehicle constraints. A centralized planner manager arbitrates among planning modules and integrates predicted object trajectories, occupancy grids, and traffic-signal states to ensure safe, lawful, and continuous navigation in complex urban environments.

\subsection{Control}
Vehicle control is realized through a combined lateral and longitudinal control architecture operating in closed loop with the planner. Lateral control is implemented using a linear Model Predictive Control (MPC) formulation that minimizes path-tracking error under steering and stability constraints by solving a constrained quadratic program at each control cycle.

Longitudinal control employs a feed-forward and feedback structure with PID compensation to regulate vehicle speed relative to the planned velocity profile, incorporating delay and slope compensation for improved tracking fidelity. Together, these controllers generate smooth, stable, and physically realistic vehicle motion in simulation.

By integrating AWSIM’s high-fidelity digital twin with Autoware’s production-grade LiDAR-based perception, hierarchical planning, and MPC-based control, our testbed enables rigorous, fully reproducible end-to-end evaluation of autonomous driving pipelines under realistic urban conditions while maintaining complete experimental safety and controllability.

}